%% file: main.tex
\documentclass[11pt,a4paper]{article}
\usepackage[hyperref]{acl2020}
\usepackage{times}
\usepackage{latexsym} 
\usepackage{amssymb}
\usepackage{amsmath}
\usepackage{booktabs}
\usepackage{url} 
\usepackage{enumitem}
\usepackage{mathtools}
\usepackage{fontawesome} 
\usepackage{contour}
\usepackage{subcaption}
\usepackage{fdsymbol}

\usepackage{pifont}% http://ctan.org/pkg/pifont
\newcommand{\cmark}{\ding{51}}%
\newcommand{\xmark}{\ding{55}}%

\setitemize{noitemsep,topsep=0pt,parsep=0pt,partopsep=0pt}

\usepackage{microtype}
\usepackage{color, colortbl}

\definecolor{Gray}{gray}{0.85}
\definecolor{LightCyan}{rgb}{0.88,1,1}
\aclfinalcopy % Uncomment this line for the final submission
\def\aclpaperid{1530} %  Enter the acl Paper ID here

\title{\vspace*{-0.5in}
{{\small \hfill EMNLP-Findings'20}\\
\vspace*{.25in}}  {C}ommon{G}en: {A} {C}onstrained {T}ext {G}eneration {C}hallenge \\ for {G}enerative {C}ommonsense {R}easoning}

\author{ 
Bill Yuchen Lin\textsuperscript{{$\varheartsuit$}}\quad  Wangchunshu Zhou\textsuperscript{{$\varheartsuit$}}\quad
Ming Shen\textsuperscript{{$\varheartsuit$}} \quad 
Pei Zhou\textsuperscript{{$\varheartsuit$}} \quad 
\\ \textbf{Chandra Bhagavatula\textsuperscript{{$\spadesuit$}} \quad Yejin Choi\textsuperscript{{$\spadesuit \vardiamondsuit$}}
\quad Xiang Ren\textsuperscript{{$\varheartsuit$}}
}
\\
{{\textsuperscript{$\varheartsuit$}University of Southern California}} \quad {\textsuperscript{$\spadesuit$}Allen Institute for Artificial Intelligence} \\ 
{\textsuperscript{$\vardiamondsuit$}Paul G. Allen School of Computer Science \& Engineering, University of Washington}\\
{
\texttt{\{yuchen.lin,xiangren\}@usc.edu,\{chandrab,yejinc\}@allenai.org}} 
} 
 
\renewcommand\footnotemark{}

\begin{document} 
\maketitle

\input{0-abs.tex}

\input{0_intro.tex}

\input{1_prob.tex}

\input{2_data.tex}

\input{3_methods.tex}

\input{4_eval.tex}

	\input{5_related.tex}

\input{6_conclu.tex}

		\section*{Acknowledgements}
This research is based upon work supported in part by the Office of the Director of National Intelligence (ODNI), Intelligence Advanced Research Projects Activity (IARPA), via Contract No. 2019-19051600007, the DARPA MCS program under Contract No. N660011924033 with the United States Office Of Naval Research, the Defense Advanced Research Projects Agency with award W911NF-19-20271, and NSF SMA 18-29268. 
The views and conclusions contained herein are those of the authors and should not be interpreted as necessarily representing the official policies, either expressed or implied, of ODNI, IARPA, or the U.S. Government. %optional
We would like to thank all the collaborators in USC INK research lab for their constructive feedback on the work.

	\bibliography{acl2020}
	\bibliographystyle{acl_natbib}
	
	\clearpage
	\input{appendix}

\end{document}

%% file: 0-abs.tex
\begin{abstract}
%Given a set of common concepts
%like ``\{\textit{apple (noun), pick (verb), tree (noun)}\}'', 
%humans find it easy to write a sentence describing a grammatical and logically coherent scenario that covers these concepts, 
% for example: {``a boy \textit{picks} an \textit{apple} from a \textit{tree}''}. 
%The process of generating these sentences requires humans to use commonsense knowledge.
%We denote this ability as a kind of \textit{generative commonsense reasoning}.
%Recent work in commonsense reasoning  has focused mainly on \textit{discriminating} the most plausible scenes from distractors via natural language understanding (NLU) settings.

% \chandra{It is unclear what is meant by "create everyday scenario".}
%  \chandra{minor thought: since "generative commonsense reasoning" is not an established term, but something we want to coin, I wonder if it might be better to introduce it later after describing the task in simpler terms.}
 
% Commonsense reasoning is a long-standing challenge for artificial intelligence. 
Recently, large-scale pretrained language models have demonstrated impressive performance on several commonsense-reasoning benchmark datasets. However, building machines with commonsense to compose realistically plausible sentences remains challenging.
%  that how we can make machines generate sentences.
% primarily due to the lack of a specialized benchmark dataset. 
In this paper, we present a constrained text generation task, \textsc{CommonGen} associated with a benchmark dataset, to explicitly test machines for the ability of \textit{generative commonsense reasoning}.
Given a set of common concepts (e.g., \{dog, frisbee, catch, throw\}); the task is to generate a coherent sentence describing an everyday scenario using these concepts (e.g., ``a man throws a frisbee and his dog catches it'').

The \textsc{CommonGen} task is challenging because it inherently requires 1) \textit{relational reasoning} with background  commonsense knowledge, and 2) \textit{compositional generalization} ability to work on unseen concept combinations.
% \chandra{a reviewer unfamiliar with the task might not know what is meant by "concept compounds".}
Our dataset, constructed through a combination of crowdsourced and existing caption corpora, consists of 
%30k concept-sets  and 50k sentences .
79k commonsense descriptions over 35k unique concept-sets.
% \yejin{can we say 30k "unique" concept-sets?}
%The crowd-workers were also asked to write the rationales (i.e. the commonsense facts) for the development and test sets.
%We conduct experiments on a variety of generation models with both automatic and human evaluation.
Experiments show that there is  a large gap between state-of-the-art text generation models (e.g., T5) and human performance.
Furthermore, we demonstrate that the learned generative commonsense reasoning capability can be transferred to improve downstream tasks 
% such as CommonsenseQA
% (76.9\% to 78.4 in dev accuracy) 
by generating additional context.
% We hope that this task facilitates future research in incorporating commonsense knowledge into natural language generation systems. 
% The models struggle at the task, often generating grammatically sound yet realistically implausible sentences -- pointing to interesting future research. 
% \chandra{need a better last sentence}
\end{abstract}

%% file: 0_intro.tex
\section{Introduction}
\label{sec:intro}

\begin{figure}[ht!]
	\centering
	\includegraphics[width=1.\linewidth]{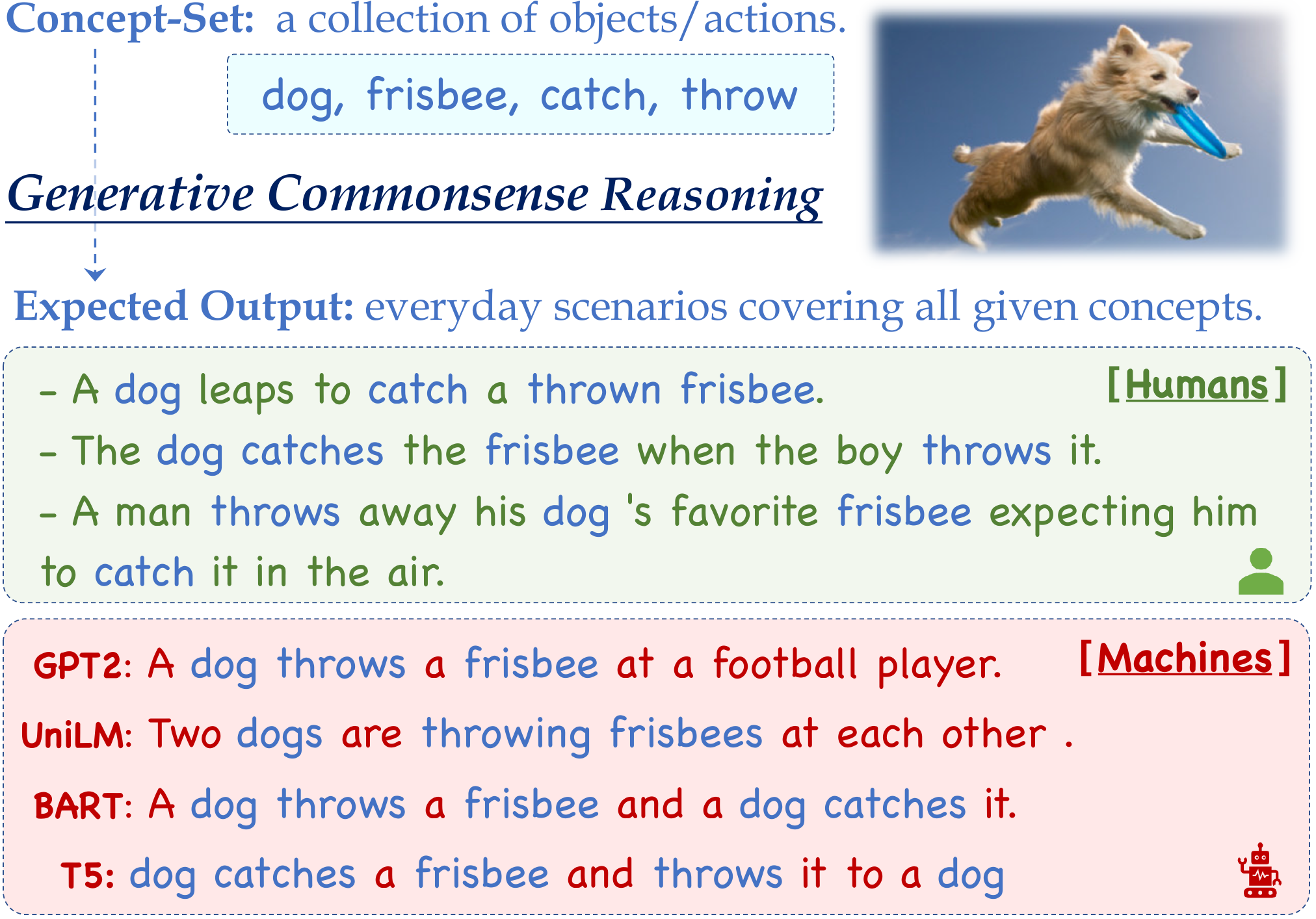}
	\caption{\small{\textbf{An example of the dataset of \textsc{CommonGen}.} GPT-2, UniLM, BART and T5 are large pre-trained text generation models, \textit{fine-tuned} on the proposed task.}
	\vspace{-1em}
	}
	\label{fig:intro}
\end{figure}

Commonsense reasoning, 
the ability to make acceptable and logical assumptions about ordinary scenes in our daily life,
has long been acknowledged as a critical bottleneck of artificial intelligence and natural language processing~\cite{davis2015commonsense}.
%A distinct characteristic of commonsense reasoning problems is that they are trivial for humans but surprisingly challenging for machine models.
Most recent commonsense reasoning challenges, such as CommonsenseQA~\cite{Talmor2018CommonsenseQAAQ}, SocialIQA~\cite{sap-etal-2019-social}, WinoGrande~\cite{Sakaguchi2019WINOGRANDEAA} and HellaSwag~\cite{Zellers2019HellaSwagCA}, have been framed as \textit{discriminative} tasks -- i.e. AI systems are required to \textit{choose} the correct option from a set of choices based on a given context. 
While significant progress has been made on these discriminative tasks, 
we argue that commonsense reasoning in text generation poses a distinct complementary challenge.
% \chandra{TODO: prev sentence needs rephrasing} 
In this paper, we advance machine commonsense towards \textit{generative} reasoning ability.
%compare the plausibility of multiple scenarios
%(constructed by combining each answer choice with the question's context) 
%and 

	\begin{figure*}[h!]
		\centering
		\includegraphics[width=1\linewidth]{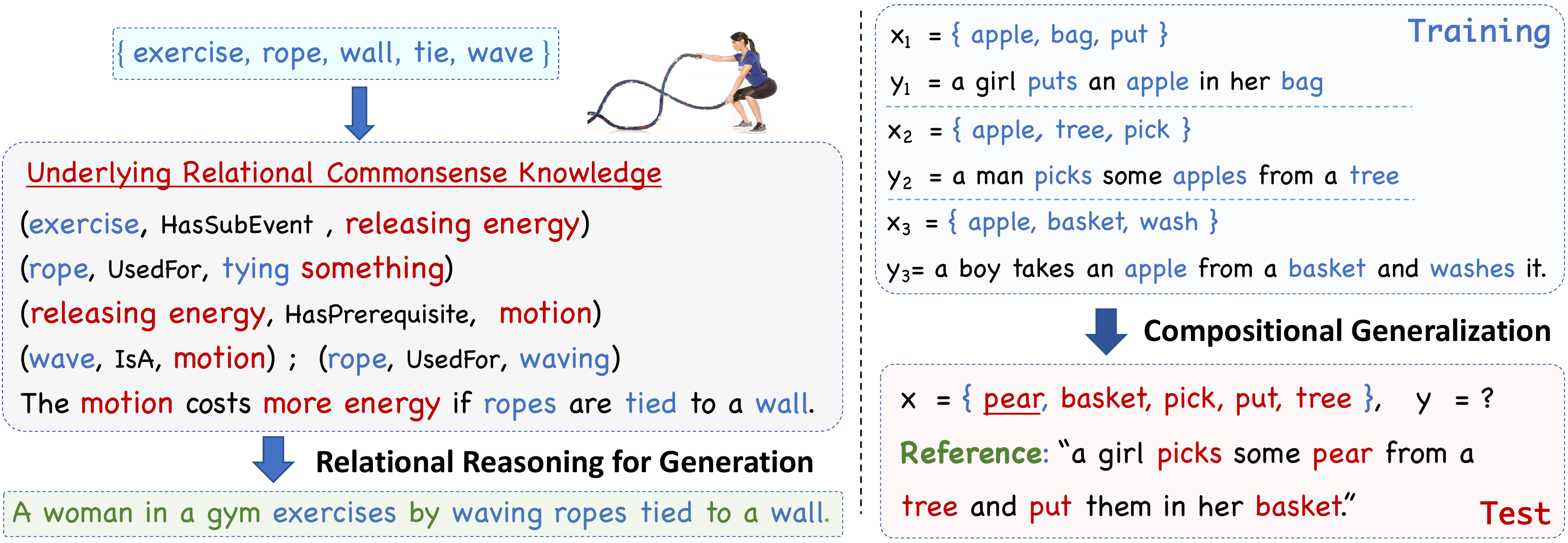}
		\caption{\small{Two \textbf{key challenges} of \textsc{CommonGen}:  \textit{relational reasoning} with underlying commonsense knowledge about given concepts (left), and \textit{compositional generalization} for  unseen combinations of concepts (right).}}
		\label{fig:challenges}
	\end{figure*}
Humans acquire the ability to compose sentences by learning to understand and use common concepts that they recognize in their surrounding environment~\cite{tincoff1999some}. The acquisition of such an ability is regarded as a significant milestone of human development~\cite{moore2013development}.  \textit{Can machines acquire such generative commonsense reasoning ability?} 
To initiate the investigation, we present \textsc{CommonGen}\footnote{\url{http://inklab.usc.edu/CommonGen/}.} -- a novel constrained generation task that requires machines to generate a sentence describing a day-to-day scene using concepts from a given \textit{concept-set}. 
% For example, in Figure~\ref{fig:intro}, given a set of concepts: \{\textit{exercise}, \textit{rope}, \textit{wall}, \textit{tie}, \textit{wave}\}, machines are required to generate a sentence such as ``a man in a {gym} \textit{exercises} by \textit{waving} \textit{ropes} \textit{tied} to a \textit{wall}.''
For example, in Figure~\ref{fig:intro}, given a set of concepts: \{\textit{dog}, \textit{frisbee}, \textit{catch}, \textit{throw}\}, machines are required to generate a sentence such as ``a man \textit{throws} a {frisbee} and his \textit{dog} \textit{catches} it in the air.''

To successfully solve the task, models need to incorporate two key capabilities: a) \textit{relational reasoning}, and b) \textit{compositional generalization}. 
Grammatically sound sentences may not always be realistic as they might violate our commonsense (e.g., \textit{``a dog throws a frisbee ..."}). In order to compose a plausible sentence that describes an everyday scenario, models need to construct a grammatical sentence while adhering to and reasoning over the commonsense relations between the given concepts.
% Through analysis, we find that our dataset requires complex compositions of various types of commonsense knowledge such as spatial relations, object properties, social conventions, temporal events, etc.
Models additionally need \textit{compositional generalization} ability to infer about unseen concept compounds. This encourages models to reason about a potentially infinite number of novel combinations of familiar concepts -- an ability believed to be a limitation of current AI systems~\cite{lake2018generalization, keysers2020measuring}.

Therefore, in support of the \textsc{CommonGen} task, we present a dataset consisting of 35,141 concept-sets associated with 77,449 sentences.
% \yejin{these numbers might be out-dated? or they don't seem to add up from the table later...} 
We explicitly design our dataset collection process to capture the key challenges of relational reasoning and compositional generalization described above, through an actively controlled crowd-sourcing process.
We establish comprehensive baseline performance for state-of-the-art language generation models with both extensive automatic evaluation and manual comparisons. The best model, based on T5 \cite{raffel2019exploring}, achieves 28.86\% with significant gap compared to human performance of 52.43\% in the \texttt{SPICE} metric -- demonstrating the difficulty of the task.
Our analysis shows that state-of-the-art models struggle at the task, generating implausible sentences -- e.g. ``dog throws a frisbee ..." , ``giving massage to a table", etc. 
Additionally,
we show that successful \textsc{CommonGen} models can benefit downstream tasks (e.g., commonsense-centric question answering) via generating useful context as background scenarios.
We believe these findings point to interesting future research directions for the community of commonsense reasoning. 
% We hope \textsc{CommonGen} will be a beneficial resource for both commonsense reasoning and language generation communities. 
% \chandra{concrete numbers in this paragraph will be helpful.}
%We hope \textsc{CommonGen} could be an initial resource for future 

% The experiments show that all methods are far from human performance in generative commonsense reasoning.
%	Interestingly, we find that models trained with distantly-supervised data can perform surprisingly good in other deterministic reasoning tasks like CSQA.

% The major contributions of this work are summarized as follows: 
% 1) We introduce the first large-scale constrained text generation dataset targeting at generative commonsense reasoning;
% 2) We systematically compare methods for this constrained text generation with extensive experiments and evaluation. 
% 3) Our code and data are publicly available, so future research in this direction can be directly developed in a unified framework. 
%		\item We show that the it is promising to transfer distantly-supervised models to other downstream tasks without further annotations.
%	\end{itemize}

%% file: 1_prob.tex
	\section{Task Formulation and Key Challenges}
	\label{sec:formualtion}
	We formulate the proposed \textsc{CommonGen} task with mathematical notations and discuss its inherent challenges with concrete examples. 
	The input is an unordered set of $k$ concepts $x=\{c_1,c_2,\dots,c_k\}\in \mathcal{X}$ (i.e. a concept-set), where each concept $c_i\in \mathcal{C}$ is a common object (noun) or action (verb). 
	We use $\mathcal{X}$ to denote the space of all possible concept-sets and use $\mathcal{C}$ to denote the concept vocabulary (a subset of ConceptNet's unigram concepts).
	The expected output is a simple, grammatical sentence $y\in\mathcal{Y}$ that describes a common scenario in our daily life, using  all given concepts in $x$ (morphological inflections are allowed).
	A scenario can depict either a static situation or a short series of actions.
% 	In addition, we also provide rationales as an optional resource to model the generation process.
% 	For each pair of $(x, y)$, a rationale $r$ is a list of sentences that explains the background commonsense knowledge used in the scenario recovering process (see the gray box in Fig.~\ref{fig:intro}). 
	The \textsc{CommonGen} task is to learn a function $f:
	\mathcal{X} \rightarrow \mathcal{Y}$, which maps a concept-set $x$ to a sentence $y$. 
%	Thus, it can be seen as an instance of \textit{constrained text generation}~\cite{Hu2017TowardCG}.
	The unique challenges of this task come from two aspects:
	
%	\smallskip
%	\noindent
%	\textbf{Constrained Sentence Decoding.}
%	Lexically constrained sentence decoding has been studied in the machine translation community~\cite{Hokamp2017LexicallyCD}. They mainly focus on the situation when specific alignment of words or phrases (e.g. terminology) must present in target sentences. 
%	Simply ordering a bag of words for recovering a complete sentence based on syntactic information~\cite{Zhang2015DiscriminativeSW} has also been investigated.
%	However, it is still an open problem how to generate sentences given an \textit{unordered} set of \textit{multiple} keywords with potential \textit{morphological changes} (e.g. ``pick'' $\rightarrow$ ``picks'').
%	Apart from that, the part-of-speech constraints also brings more difficulties (e.g. ``place'' can be verb/noun).
%	
	
	\smallskip
	\noindent
	\textbf{Relational Reasoning with Commonsense.}
	Expected generative reasoners should prioritize the most plausible scenarios over many other less realistic ones.
	As shown in Figure~\ref{fig:challenges},
	models need to recall necessary relational commonsense facts that are relevant to the given concepts, and then reason an optimal composition of them for generating a desired sentence.
% 	Recall the first illustrative examples in Figure~\ref{fig:intro},
%     the underlying knowledge are implicit and compositional: (a) dogs love to perform tricks with humans, (b) catching a frisbee is a trick and (c) humans love to play this game with dogs.
%     As for the other example in Section~\ref{sec:intro} about \{\textit{exercise}, \textit{rope}, \textit{wall}, \textit{tie}, \textit{wave}\}, we also need to compose the following commonsense facts: (i) doing exercises is to cost energy, (ii) waving a rope can cost energy, and (iii) it is more useful when the rope is tied to a wall.
	In order to complete a scenario,  generative commonsense reasoners also need to reasonably associate additional concepts (e.g., `woman',  `gym') as agents or background environments for completing a coherent scenario. 
%	 such as \textit{``a boy picks {an apple tree} and places it into bags''} or \textit{``a boy {places some bags on a tree} and picks an apple''}. 

	This not only requires understanding underlying commonsense relations between concepts, 
	but also incrementally composing them towards a globally optimal scenario.
	The underlying reasoning chains are  
	inherently based on a variety of background knowledge such as spatial relations,  object properties, physical rules, temporal event knowledge, social conventions, etc. 
	However, they may not be recorded in any existing knowledge bases.
% 	and thus the task can be also viewed as learning to querying them from textual corpora or large pre-trained language models  towards completing an everyday scenario.

% 	Recall the illustrative example in Figure~\ref{fig:intro},
% 	even such a simple scenario generation process needs pretty much commonsense knowledge like: 1) ``\textit{apples grow in trees}''; 2) ``\textit{bags are containers that you can put something in}''; 3) ``\textit{you usually pick something and then place it in a container}''.
	% able to model the distribution $P(\texttt{scene}~|~c_1,c_2,\dots, c_n)$ for selectively generate the most plausible scene

	\smallskip
	\noindent
	\textbf{Compositional Generalization.}
% 	Testing concept-sets usually contain rare concept co-occurrence or even unseen concepts.
	Humans can compose a sentence to describe a scenario about the concepts they may never seen them co-occurring.
	For example, in Figure~\ref{fig:challenges}, there is a testing concept-set $\hat{x}=$\{\textit{pear}, \textit{basket}, \textit{pick}, \textit{put}, \textit{tree}\}.
	The concept `pear' never appear in the training data, and `pick' never co-occurs with `basket'. 
% Meanwhile, there are some relevant training examples: 
% \begin{itemize}
%     \item $x_1=$\{\textit{apple}, \textit{bag},  \textit{put}\} $\rightarrow y_1=$``a boy \textit{puts} an \textit{apple} in a \textit{bag}.''
%     \item $x_2=$\{\textit{apple}, \textit{tree},  \textit{pick}\} $\rightarrow y_2=$``a girl \textit{picks} an \textit{apple} from the \textit{tree}.''
%     \item  $x_3=$\{\textit{apple}, \textit{basket},  \textit{wash}\} $\rightarrow y_3=$``a girl takes an \textit{apple} from the \textit{basket} and \textit{washes} it.''
% \end{itemize}
We, humans, can generalize from these seen scenarios in the training data and infer that a plausible output: $\hat{y}=$\textit{``a girl picks some pears from a tree and put them into her basket.''} 
This compositionally generalization ability via analogy, i.e., to make ``infinite use of finite means''~\cite{chomsky1965aspects}, is challenging for machines.
This analogical challenge not only requires inference about similar concepts (e.g., `apple' $\rightarrow$ `pear') but also their latent associations.
% to reason about complex, unseen composition by analogy,

%% file: 2_data.tex
	\section{Dataset Construction and Analysis}
	\label{sec:dataset}
\begin{figure}[t!]
		\centering
		\includegraphics[width=0.9\linewidth]{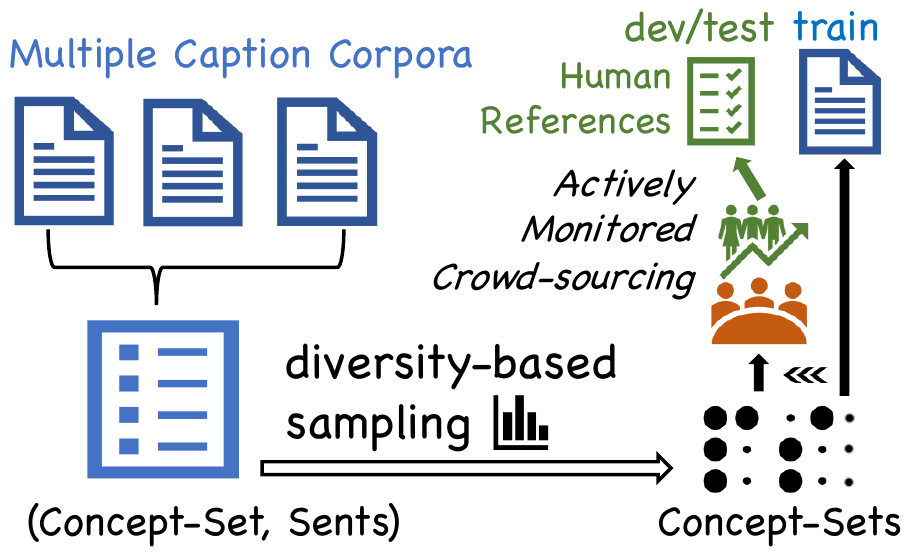}
		\caption{ \textbf{Dataset construction workflow overview.} 
		\vspace{-1em}
		}
		\label{fig:builddata}
	\end{figure}

% 	We now introduce the construction and analysis of the dataset for the proposed \textsc{CommonGen} task.
% 	To ensure that the concepts in each input concept-set are likely to be present together in a everyday scene, 
Figure~\ref{fig:builddata}
illustrates the overall workflow of our data construction for the proposed \textsc{CommonGen} task.	
We utilize several existing caption corpora for sampling frequent concept-sets (Sec.~\ref{ssec:conceptset}) for reflecting common scenarios.
	We employ AMT crowd workers for collecting human-written sentences (Sec.~\ref{ssec:amt}) for the development and test set, while we carefully monitor the quality of crowd workers and refine them dynamically.
	Finally, we present the statistics of the \textsc{CommonGen} dataset, and the analysis on the challenges  (Sec.~\ref{ssec:analysis}).

% 	1) {Concept-set collection.} 
% 	We first collect a large amount of high-quality image/video caption sentences from several existing corpora, 2) \textbf{} Then, we compute co-occurrence statistics about concept-sets of different sizes ($3\sim5$), such that we can find the concept-sets that are more likely to be present in the same scene.
% 	3) Finally, we ask crowd-workers from AMT to write scenes with rationales for every given concept-set, which serve as our development and test sets. 
% 	The training set consists of carefully post-processed human-written caption sentences, which have little overlap with dev/test sets.
% 	We present the statistics and show its inherent challenges at the end of this section.
%     \yuchen{Add a simple illustrative figure to show the data collection phase.}	

	\subsection{Collecting Concept-Sets from Captions}
	\label{ssec:conceptset}
	% We refer the combination of these caption sentences together as the caption corpus for simplicity.
% 	Following the general definition in the largest commonsense knowledge graph, ConceptNet~\cite{Speer2017ConceptNet5A}, we understand a concept as a common noun or verb.
% 	We aim to test the ability of generating natural scenes with a given set of concepts.
%	These concept-sets in our data should have high probability of being present together, otherwise generating scenes of them is hardly meaningful.
%	We assume if a set concepts are all mentioned together in more caption sentences, then this set is more plausible to be present.
    %It is obviously nonsense if we ask a reasoner to generate a scenario about an arbitrarily concept-set, which is  impossible even for humans. 
	%The expected concept-sets of our task are supposed to be very likely to co-occur in common, daily-life scenes.
	It can be unreasonable to present any \emph{arbitrary} set of concepts (e.g., $x=$\{\textit{apple, fold, rope}\}) and ask a reasoner to generate a commonsense scenario, since such an arbitrary set of concepts can be too unrelated. Therefore, our concept-sets are supposed to reflect reasonable concept co-occurrences in everyday situations.
    %These everyday scenarios are ubiquitous in images and video clips, and this leads us to consider using image and video captioning datasets as a natural resource for collecting concept-sets and sentences.
    As web images and video clips capture diverse everyday scenarios, 
    we use their caption text as a natural resource for collecting concept-sets and their corresponding descriptions of commonsense scenarios.
% 	The concepts in images/videos captions, which usually describe scenes in our daily life, thus possess the desired property.
%
	More specifically, we collect visually-grounded sentences from several existing caption datasets, including {image captioning} datasets, such as  {Flickr30k}~\cite{young-etal-2014-image}, {MSCOCO}~\cite{Lin2014MicrosoftCC}, {Conceptual Captions}~\cite{Sharma2018ConceptualCA}, as well as {video captioning} datasets including {LSMDC}~\cite{lsmdc}, {ActivityNet}~\cite{krishna2017dense}, and {VATEX}~\cite{Wang_2019_ICCV}.
	
% 	We assume if a set of concepts are all mentioned together in more caption sentences, then this concept-set is more like to co-occur.
	We first conduct part-of-speech tagging over all sentences in the corpora such that words in sentences can be matched to the concept vocabulary of ConceptNet.
	Then, we compute the sentence frequency of 
	concept-sets consisting of  3$\sim$5 concepts.
	That is, for each combination of three/four/five concepts in the vocabulary, we know how many sentences are in the corpora covering all concepts.
	
	%Towards building a more representative dataset, we expect our selected subset of concept-sets can reflect the distribution in the real world.
	Ideally, we want the selected concept-sets in our dataset to reflect the natural distribution of concept-sets in the real world. 
	%A straightforward intuition is to directly treat the frequency as the measure for likelihood of concept-sets, and then conduct probabilistic sampling based on this distribution. 
	At first glance, a reasonable solution may seem to sample from the distribution of the concept-sets based on their frequencies in the source datasets. 
	%However, this method tends to sample concept-sets that contain one or two single highly frequent concept, thus leading to corpus-dependent bias. 
	However, we find that this method leads to a rather unnaturally skewed collection of concept-sets, due to the inherent data biases from the source datasets.   
% 	Also, merely using the sentence number can be imprecise to measure the scenario diversity since many images and videos were sampled interdependently when those corpora were built.\yejin{Not sure if I understand the previous sentence and whether the point that the previous sentence is trying to make is overly important...}
	We therefore design a function to score a concept-set $x$ based on  \textit{scene diversity} and  \textit{inverse frequency penalty}.
	We denote $S(x)$ as the set of unique sentences that contain all given concepts $\{c_1, c_2, \dots, c_k\}$, and then we have  
	{{$$ \texttt{score}(x) =|S(x)|   \frac{|\bigcup_{s_i \in S(x)}\{w | w \in s_i\}|}{\sum_{s_i \in S(x)} \text{len}(s_i)} \rho(x),$$}}where $\rho(x) = \frac{|\mathcal{X}|}{\max_{c_i\in x}|\{x' ~|~ c_i \in x' ~~\text{and}~~ x' \in \mathcal{X}\}|}.$
	The first term in $\texttt{score}$ is the number of unique sentences covering all given concepts in $x$, and
	the second term is to represent the diversity of the scenes described in these sentences.
	Th last term $\rho(x)$ is the penalty of inverse frequency.
	Specifically, we find the concept in $x$ that has the maximum ``set frequency'' (i.e., the number of unique concept-sets containing a particular concept), then we take the inverse with the number of all concept-sets for normalization. 
	This penalty based on inverse set-frequency effectively controls the bias towards highly frequent concepts.
	With the distribution of such scores of concept-sets, we sample our candidate examples for the next steps.
% 	Each concept-set is associated with at least one caption sentences.
% 	We carefully post-process them and take the shortest ones with minimal overlaps as the final data.
% 	These initial concept-sets are further divided into three parts: train/dev/test.
	% we filter the concept-sets in training data such that each dev/test concept-sets has at least three unseen concepts.
	% Simply put, 
% 	We then iterate all training concept-sets and remove the ones that have more than two overlapping concepts with any concept-set in the dev or test set.
% 	Thus, the dev/test set can better measure the generalization ability of models on unseen combinations of concepts.
	 
	\begin{table}[t]
	\small
		\centering
		\scalebox{0.94
		}{
			\begin{tabular}{@{}l|ccc@{}}
				\toprule
			\textbf{Statistics} & \textbf{Train} & \textbf{Dev} & \textbf{Test} \\ \midrule
        \textbf{\# Concept-Sets} & \textbf{32,651} & \textbf{993} & \textbf{1,497} \\
        \qquad {-Size} = 3 & 25,020 & 493 & - \\
        \qquad {-Size} = 4 & 4,240 & 250 & 747 \\
        \qquad {-Size} = 5 & 3,391 & 250 & 750 \\
        
        \midrule
        {\textbf{\# Sentences }} & {67,389} & {4,018} & {7,644} \\ 
        {\textbf{per Concept-Set }} & 2.06 & 4.04 & \textbf{5.11} \\ 
        {\textbf{Average Length}} & 10.54 & 11.55 & 13.28 \\ 
        \midrule 
        \midrule 
        \textbf{\# Unique Concepts} & 4,697 & 766 & 1,248 \\
        \textbf{\# Unique Concept-Pairs } & 59,125 & 3,926 & 8,777 \\
        \textbf{\# Unique Concept-Triples } & 50,713 & 3,766 & 9,920 \\
        \midrule
        % \rowcolor{LightCyan}
        \textbf{\% Unseen Concepts} & - & 6.53\% & 8.97\% \\
        % \rowcolor{LightCyan}
        \textbf{\% Unseen Concept-Pairs} & - & 96.31\% & 100.00\% \\
        % \rowcolor{LightCyan}
        \textbf{\% Unseen Concept-Triples} & - & 99.60\% & 100.00\% \\
        
				\bottomrule                        
			\end{tabular}} 
		\caption{The \textbf{basic statistics} of the \textsc{CommonGen} data. We highlight the ratios of concept compositions that are unseen in training data, which assures the challenge in compositional generalization ability.}
		\label{tab:basicstat}
	\end{table}

	\subsection{Crowd-Sourcing References via AMT}
	\label{ssec:amt}
% 	\yejin{How about restructuring this subsection to start right away to declare that in order to ensure the best quality, all reference captions in the dev and test sets are crowdsourced, which amounts to 10,000 reference captions over 2500 distinct concept-sets (to better highlight the larger number 10,000, not just the total number of concept-sets being 2500). and then for the specific reasons for quality control, we could also mention the potential concerns for data-leak (e.g., new SOTA language models might have seen the source text of the caption datasets during pre-training). at that point, i think it'd be fine to drop the mention of "too expensive cost of human efforts and the difficulty in verifying the quality of connected sentences" as the former is unnecessary and the later is rather self-shooting... }
    In order to ensure the best quality,
     the references of the evaluation examples are crowdsourced from crowd workers on \textit{Amazon Mechanical Turk},
     which amounts to \textbf{10,060} references over 2.5k distinct concept-sets.
     Note that these newly collected references for dev and test examples can ensure that we can do a fair comparisons targeting generalization, considering potential data-leak (i.e., recent pre-trained language models might have seen the caption datasets). 
    Each concept-set was assigned to at least 3  workers. 
  	In addition to references about given concept-sets, we also ask the workers to provide rationale sentences to explain what commonsense facts they have used, for ensuring that the described scenarios are common in daily life (example rationales are shown in Fig~\ref{fig:morecase}).
% 	Although the human-written sentences in the caption corpora can be seen as quality annotations for the \textsc{CommonGen} task as well, 
% 	they were written with specific visual context (i.e. an image or a video clip).
% 	Toward better diversity of the scenes about sampled concept-sets and more rigorous evaluation for systems, crowd-sourcing additional human references is necessary that are written with only concept-sets as the context.
% 	We decide to use the AMT platform for collecting such sentences for covered the top-ranked 2,500 concept-sets in the sampled results,
% 	due to the expensive cost of human efforts in writing sentences and the difficulty in verifying the quality of collected sentences.

    We control the quality by actively filtering workers who produced low-quality references, then removing their annotations, and finally re-opening the slots only for quality workers. {There were 1,492 accepted workers in total and 171 disqualified workers in the end after the active filtering.}
    There are three criteria for efficiently narrowing down candidates for us to further manually remove out low-quality workers: 1) coverage via part-of-speech tagging, 2) especially high perplexity via GPT-2, and 3) length of the rationales.
    Meanwhile, we also dynamically replaced the concept-sets that majority of the references do not make sense to ensure the final quality.
    
    \subsection{Permutation-Invariant Annotating}
    We have to present every input as a a string for annotators to read, which means we take a random permutation of the concept-set as a linear sequence. Therefore we may wonder if annotators will make flexible adjustment on the concept order when creating references for CommonGen.
	To address this concern, 
	we first study the correlation between the input concept-order and the reference concept-order (i.e., the order of the given concepts in the human annotations).
	We find that 96.97\% of the references, of which the concept-order is different from the order shown when they are annotating.
	
    More specifically, 
    we use Spearmans's rank correlation coefficient to understand the correlation between input concept-order and reference concept-order.
    It turns out that the mean correlation over all input-reference pairs on test examples is -0.031, which suggests that different permutation of the input concept-order do not have notable influence on the order of concept in the human references, thus being permutation-invariant.
	
	\subsection{Finalizing Adequate References}
	As there may be more than one acceptable scenes for each input concept-set, we would like to check if our human references are enough before we finalizing our dataset.
	Thus, we took one more round of crowd-sourcing to add one more reference for each concept-set by \textit{new} annotators.
	Then, we compute the inter-annotator agreement (IAA) by using the cosine similarity between all pairs of human references, based on SentenceBERT~\cite{reimers-gurevych-2019-sentence} (fine-tuned for semantic similarity analysis).
	Note that if we have $k$ human references for an example in the end, then we will have $k(k-1)/2$ different pairs of references, each of which has a cosine similarity between their sentence embeddings.
	Then, we take the \textit{median} of these similarity scores as a proxy to understand if we have collect adequate human references.
	
\begin{figure}[t!]
\centering
\hspace{-0.4em}
\begin{subfigure}[b]{0.47\textwidth}
   \includegraphics[width=1\linewidth]{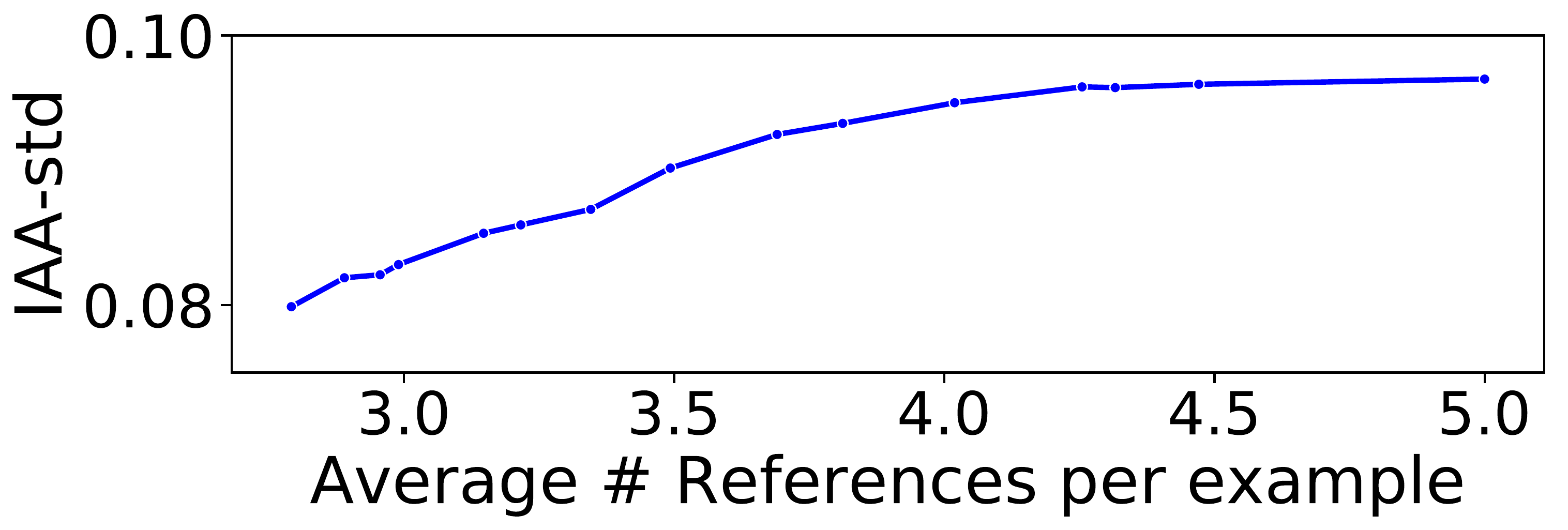}
   \label{fig:trend_std} 
\end{subfigure}

\begin{subfigure}[b]{0.482\textwidth}
    \vspace{-2.4em}
    \hspace{-0.4em}
   \includegraphics[width=1\linewidth]{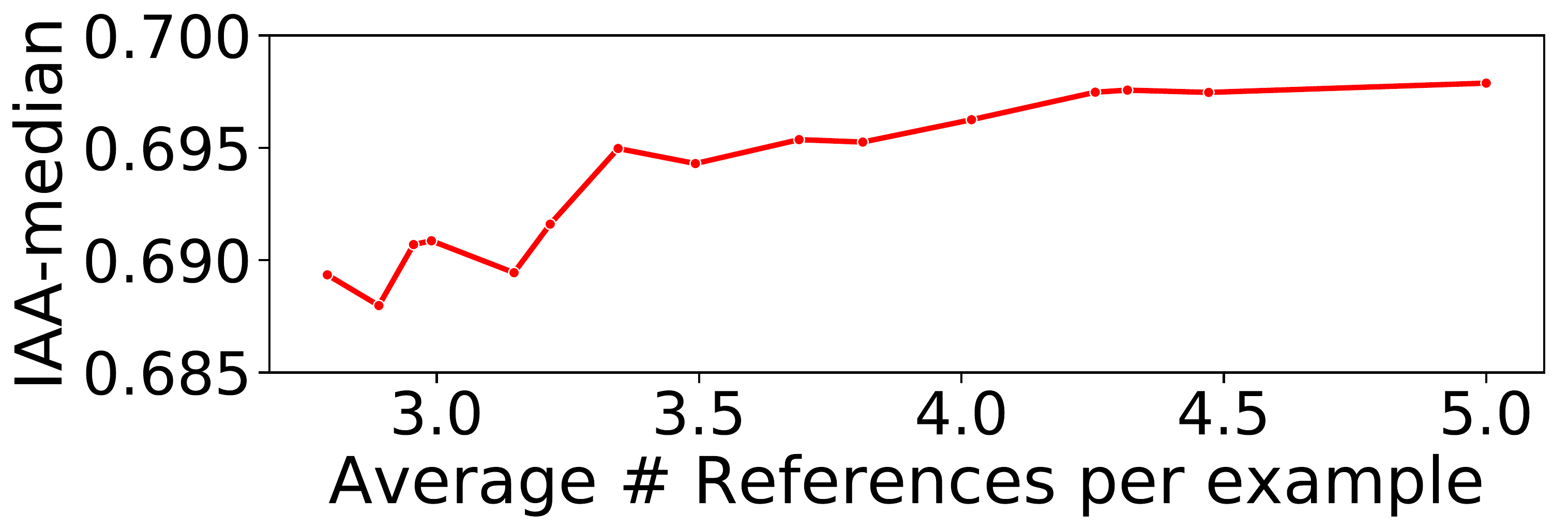}
   \label{fig:trend_median}
\end{subfigure}
\vspace{-2em}
\caption[Two numerical solutions]{The curve of inter-annotator agreement (IAA) in terms of their std (up) and median (bottom) when average number of references increase.}
\label{fig:iaacurve}
\end{figure}

	The underlying rationale here is that if there are more references that are very similar to each other yet from different annotators, then it is likely that current references are adequate for this example.
	As shown in Figure~\ref{fig:iaacurve}, we simulated different sizes of number of references per example. We find that the IAA will be saturated when we have the fifth ones, and thus we believe references are adequate. 
	Also, from the \textit{std} of these IAA scores, we find that the diversity of the references, and it also saturate when there are five references.
% 	We collect more human-written scenes for each concept-set in dev and test set through crowd-sourcing via the {Amazon Mechanical Turk} platform. 
% 	Each input concept-set is annotated by at least three different humans.
% 	The annotators are also required to give sentences as the rationales, which further encourage them to use common sense in creating their scenes.
%	With the metric we introduced in Sec.~\ref{sec:evaluation},
%	we find that the crowd-sourced scenes have high inter-annotator agreement, which means that human-beings as generative reasoners have a relatively deterministic scope of the created scenes.
% 	The crowd-sourced sentences correlate well with the associated captions, meaning that it is reasonable to use caption sentences as training data although they can be partly noisy.
% 	Additionally, we utilize a search engine over the OMCS corpus~\cite{Singh2002OpenMC} for retrieving relevant propositions as distant rationales in training data. 

% \begin{figure*}
% 	\centering
% 	\includegraphics[width=1\linewidth]{test_freq}
% 	\caption{The frequency of top 50 single concepts (upper) and co-occurred concept-pairs (lower) in the test data. }
% 	\label{fig:conceptcoocur}
% \end{figure*}

% 	\subsection{Statistics}
% 	\label{ssec:stat}
    % We use these 2,500 concept-sets as the dev and test set examples for their higher weights and better diversity of human-written sentences.

    % We sample from the remaining concept-sets as the training examples, for which we use the associated captions as the target outputs.
    % Note that we e

	\subsection{Down-Sampling Training Examples }
	
    In order to evaluate the \textit{compositional generalization} ability, 
    we down-sample the remaining candidate concept-sets to construct a distantly supervised training dataset (i.e., using caption sentences as the human references). 
    We explicitly control the overlap of the concept-sets between training examples and dev and test examples.
	The {basic statistics }of the final dataset is shown in Table~\ref{tab:basicstat}.
	There are on average four sentences for each example in dev and test sets, which provide a richer and more diverse test-bed for automatic and manual evaluation.
	Table~\ref{tab:basicstat} also shows the ratio of \textit{unseen concept compositions} (i.e., concept, concept-pair, and concept-triple) in the dev and test. %, which are not present in the training data. 
% 	This makes \textsc{CommonGen} challenging in terms of \textbf{compositional generalization} ability.
	Notably, all pairs of concepts in every test concept-set are unseen in training data and thus pose a challenge for compositional generalization.

%     \begin{figure}[t!]
% 		\centering
% 		\includegraphics[width=1\linewidth]{trend.pdf}
% 		\caption{\small \textbf{Inter-Annotator agreement} (median of cosine similarities) over different sizes of references per example.}
% 		\label{fig:trend}
% 	\end{figure}

    \begin{figure}[t!]
		\centering
		\includegraphics[width=0.98\linewidth]{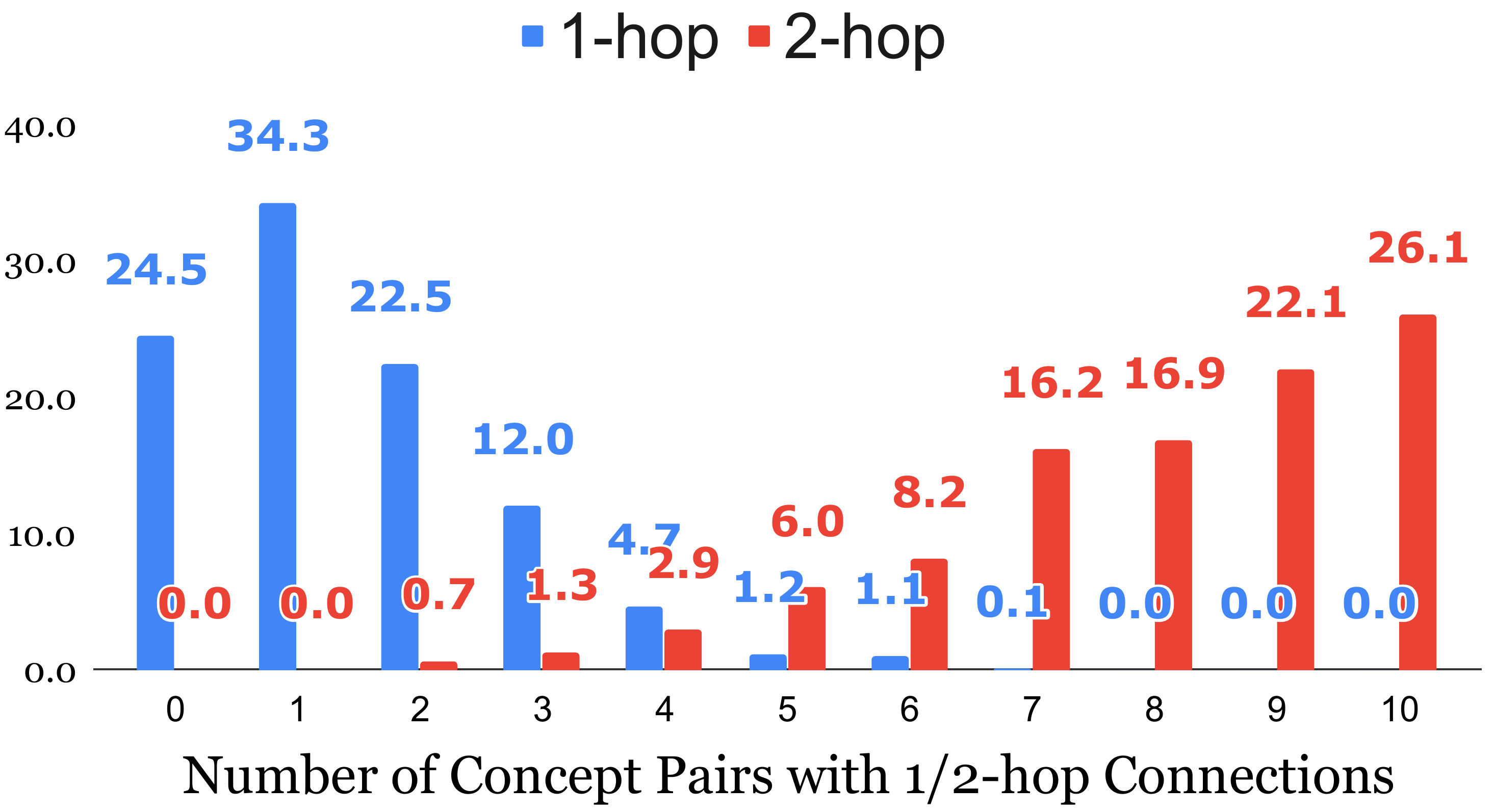}
		\caption{\small \textbf{Connectivity analysis} in 5-size concept-sets in the test set, each of which consists of 10 concept pairs. {For example, 12.0 in blue means: there are 12\% concept-sets that have 3 concept pairs with one-hop connections on ConceptNet.}}
		\label{fig:connectivity}
	\end{figure}

	\subsection{Analysis of Underlying Common Sense}
	\label{ssec:analysis}
% 	We also visualize the frequency distribution of our test concept-sets in Figure~\ref{fig:conceptcoocur} by showing the frequency of top 50 single concepts and co-occurred concept pairs.
    We here introduce deeper analysis of the dataset by utilizing the largest commonsense knowledge graph (KG), ConceptNet~\cite{Speer2017ConceptNet5A}, as an tool to study connectivity and relation types.
    
    \smallskip
    \noindent
    \textbf{Connectivity Distribution.~} 
    If the concepts inside a given concept-set is more densely connected with each other on the KG, then it is likely to be easier to write a scenario about them.
    % A direct criteria for understanding the difficulty of a given concept-set is the connectivity among them through one/two-hop links on a well-established knowledge graph like ConceptNet.
    % For example, in a 5-size concept-set, we have 10 unique pairs of concepts, each of which may or may not have one/two-hop links on the ConceptNet.
    In each 5-size concept-set (i.e. a concept-set consists of five concepts), 
    there are 10 unique pairs of concepts, the connections of which we are interested in. 
    As shown in Figure~\ref{fig:connectivity}, if we look at the one-hop links on the KG, about 60\% of the 5-size concept-set have less than one link among all concept-pairs.
    On the other hand, if we consider two-hop links, then nearly 50\% of them are almost fully connected (i.e. each pair of concepts has connections).
    These two observations together suggest that the \textsc{CommonGen} has a reasonable difficulty: the concepts are not too distant or too close, and thus the inputs are neither too difficult nor too trivial. 
    
\begin{figure}[!t]
	\centering
	\includegraphics[width=1\linewidth]{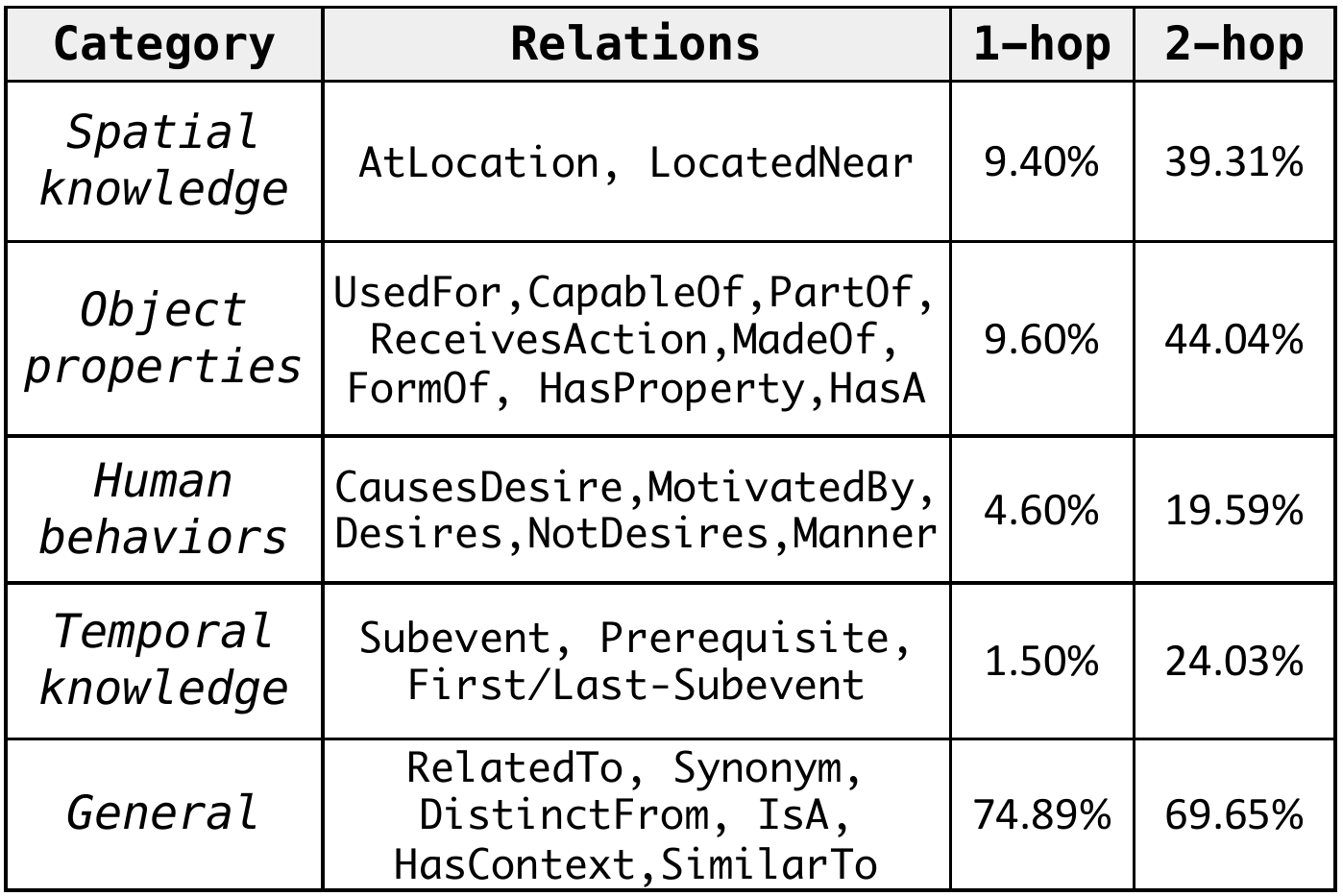}
	\captionof{table}{The \textbf{distributions of the relation categories} on one/two-hop connections.}
	\label{tab:cate}
\end{figure}

    \smallskip
    \noindent
    \textbf{Relation Distribution.~}
    Furthermore, the relation types of such connections can also tell us what kinds of commonsense knowledge are potentially useful for relational reasoning towards generation.
    We report the frequency of different relation types\footnote{ Relation definitions are at \url{https://github.com/commonsense/conceptnet5/wiki/Relations}.} of the one/two-hop connections among concept-pairs in the dev and test examples in Fig.~\ref{fig:relationdist}.
    To better summarize the distributions, 
    we categorize these relations into five major types and present their distribution in Table~\ref{tab:cate}, respectively for one/two-hop connections between concept pairs.
    
%% file: 3_methods.tex
\section{Methods}
\label{sec:baseline}
We briefly introduce the  baseline methods that are tested on the \textsc{CommonGen} task.
% As there is no principled approach for the proposed setting, to the best of our knowledge,
% we mainly consider it as a conditional sentence generation task that can be solved by many sequence-to-sequence frameworks. 

\smallskip
\noindent
\textbf{Encoder-Decoder Models.}
Bidirectional RNNs and Transformers~\cite{Vaswani2017AttentionIA} are two most popular architectures for seq2seq learning.
We use them with the addition of attention mechanism~\cite{Luong2015EffectiveAT} with copying ability~\cite{gu-etal-2016-incorporating}, which are based on an open-source framework OpenNMT-py~\cite{Klein2017OpenNMTOT}.
We use \texttt{bRNN-CopyNet} and \texttt{Trans-CopyNet} denote them respectively.
To alleviate the influence from the concept ordering in such sequential learning methods, 
we randomly permute them multiple times  for training and decoding and then get their average performance.
To explicitly eliminate the order-sensitivity of inputs,
we replace the encoder with a mean pooling-based MLP network (\texttt{MeanPooling-CopyNet}).
% Similarly, we consider removing the position embedding module from the Transformer as well (``Set. Trans.'').  

\smallskip
\noindent
\textbf{Non-autoregressive generation.} 
% 	including \textit{iNAT}~\cite{Lee2018DeterministicNN}, \textit{Insertion Transformer} (InsertTrans)~\cite{Stern2019InsertionTF}, and
Recent advances~\cite{Lee2018DeterministicNN, Stern2019InsertionTF} in conditional sentence generation have an emerging interest on (edit-based) non-autoregressive generation models, which iteratively refine generated sequences.
We assume that these models potentially would have better performance because of their explicit modeling on iterative refinements, and thus study the most recent such model {Levenshtein Transformer} (\texttt{LevenTrans}) by~\citeauthor{Gu2019LevenshteinT} (\citeyear{Gu2019LevenshteinT}). 
We also include a recent enhanced version, \texttt{ConstLeven}~\cite{Susanto2020LexicallyCN}, which incorporates lexical constraints in \texttt{LevenTrans}.

\begin{table*}[th!]
	\centering
	\scalebox{0.88
	}{
		\begin{tabular}{@{}c|cc|cc|c|cc|c@{}}
		
\textbf{Model $\backslash$ Metrics}             & \multicolumn{2}{c}{\texttt{ROUGE-2/L}} & \multicolumn{2}{c}{\texttt{\textbf{BLEU}-3/\textbf{4}}}  & \texttt{{METEOR}} & \texttt{\textbf{CIDEr}} & \texttt{\textbf{SPICE}} &  \texttt{Coverage}  \\ \midrule 
bRNN-CopyNet~\cite{gu-etal-2016-incorporating} & 7.67 & 27.77 & 12.58 & 7.06 & 16.38 & 5.06 & 13.39 &  51.15 \\ 
% bRNN w/ OMCS & 4.15 & 21.74 & 7.70 & 2.80 & 15.10 & 5.22 & 14.30 & 37.20 & 50.67 \\ 
% bRNN w/ order & 4.28 & 30.44 & 8.40 & 3.60 & 18.20 & 6.68 & 14.70 & 53.20 & 67.16 & 7.19 \\  
% Set-Trans.~\cite{Vaswani2017AttentionIA} & 1.59 & 12.96 & 3.20 & 1.20 & 8.60 & 2.02 & 7.00 & 17.71 & 24.10 \\ 
Trans-CopyNet & 8.64 & 27.85 & 12.47 & 7.56 & 15.91 & 4.65 & 12.85 & 49.06 \\
MeanPooling-CopyNet & 9.65 & 31.15 & 12.29 & 7.08 & 17.10 & 5.18 & 15.18 & 55.70 \\ 
% Trans. w/ OMCS & 1.73 & 12.81 & 4.20 & 1.40 & 9.20 & 2.48 & 8.10 & 14.93 & 20.87  \\
% Trans. w/ order & 3.31 & 18.39 & 5.90 & 2.70 & 12.50 & 3.74 & 9.70 & 32.27 & 38.71 & 3.72 \\ 
LevenTrans.~\cite{Gu2019LevenshteinT} & 10.61 & 31.87 & 21.51 & 12.65 & 20.50 & 7.45 & 16.84 & 63.81 \\ 
ConstLeven.~\cite{Susanto2020LexicallyCN} & 11.77 & 33.04 & 20.87 & 11.26 & 25.23 & 10.80 & 20.05 & 94.51 \\ 
\midrule   
GPT-2~\cite{radford2019language} & 16.85 & 39.01 & 33.92 & 23.73 & 26.83 & 12.19 & 23.57 & 79.09  \\  
BERT-Gen~\cite{bao2020unilmv2}  & 17.78 & 40.21 & 33.29 & 23.47 & 28.25 & 12.61 & 24.82 & 86.06  \\
UniLM~\cite{Dong2019UnifiedLM} & 21.20 & \textbf{43.60} & \underline{41.82} & \underline{30.73} & 30.62 & \underline{14.89} & 27.43 & 89.19 \\ 
% UniLM w/ OMCS & 20.77 & 41.25 & 36.40 & 25.70 & 29.30 & 14.08 & 29.20 & 86.32 & 89.42 \\ 
UniLM-v2~\cite{bao2020unilmv2} & 18.11 & 40.51 & 34.31 & 24.53 & 29.04 & 13.19  & 25.52 & 89.13  \\  
BART~\cite{Lewis2019BARTDS} & \textbf{22.02} & 41.78 & 39.52 & 29.01 & \textbf{31.83} & 13.98 & \underline{28.00}  & \textbf{97.35}  \\ 
 T5-Base~\cite{raffel2019exploring} & 14.63 & 34.56 &  28.76 &  18.54 & 23.94 & 9.40 & 19.87 &  76.67  \\
 T5-Large~\cite{raffel2019exploring} & \underline{21.74} & \underline{42.75} &  \textbf{43.01} &  \textbf{31.96} & \underline{31.12} & \textbf{15.13} & \textbf{28.86} &  \underline{95.29}  \\  

\midrule
% \rowcolor{LightCyan} 
Human Performance (Upper Bound) & 36.72 & 53.45 & 52.55 & 46.49 & 38.79 & 37.64 & 52.43 & 99.33  \\ 
\bottomrule

\end{tabular}
		
	} 
	\caption{\textbf{Experimental results} of different baseline methods on the \textsc{CommonGen} test set (v1.1). The first group of models are non-pretrained models, while the second group is large pretrained models that we have fine-tuned. The best models are \textbf{bold} and second best ones are \underline{underlined} within each metric. We highlight the metrics that we used in our official leaderboard. (Results on dev set are at  Table.~\ref{tab:devexp}.)}
	\label{tab:exp}
\end{table*}

\smallskip
\noindent
\textbf{Pre-trained Language Generation Models.}
We also employ various pre-trained language generation models, including \texttt{GPT-2}~\cite{radford2019language}, \texttt{UniLM}~\cite{Dong2019UnifiedLM},
\texttt{UniLM-v2}~\cite{bao2020unilmv2},
\texttt{BERT-Gen}~\cite{bao2020unilmv2}, 
\texttt{BART}~\cite{Lewis2019BARTDS}, and \texttt{T5}~\cite{raffel2019exploring}, to tackle this task and test their generative commonsense reasoning ability.
We fine-tuned all the above models on our training data with a seq2seq format.

Specifically, to use \texttt{GPT-2} for this sequence-to-sequence task, we condition the language model
on the format ``\textit{$c_1~c_2~\dots~c_k = y$}'' during fine-tuning, where $c_i$ is a concept in the given concept-set and connects with other concepts with a blank; $y$ is a target sentence.
For inference, we sample from the fine-tuned
\texttt{GPT-2} model after a prompt of ``\textit{$c_1~c_2~\dots~c_k =$}'' with beam search and use the first generated
sentence as the output sentence. 
For \texttt{BERT-Gen}, we use the \texttt{s2s-ft} package\footnote{\small{\url{https://github.com/microsoft/unilm}}} to fine-tune them in a sequence-to-sequence fashion that is similar to the LM objective employed by UniLM.

As for \texttt{T5}, the state-of-the-art text-to-text pre-trained model which is pre-trained with a multi-task objective by prepending a task description before the input text, we prepend the input concept set with a simple prompt: ``\texttt{generate a sentence with:}'' and fine-tune the model with the source sentence on the format ``\textit{generate a sentence with $c_1~c_2~\dots~c_k$}.'' For decoding, we employ the standard beam search with a beam size of 5 for all compared models.
We also report their results with a lexically-constrained decoding method, dynamic beam allocation (DBA)~\cite{post-vilar-2018-fast}, which do not show improvement over conventional beam searching.
\footnote{The used hyper-parameters are reported in the appendix. }

%into the  novel unified pre-trained language generation model, named UniLM, which uses pre-trained BERT as the encoder and then fine-tunes its whole architecture with many different generation objective. 
%It achieved the \textit{state-of-the-art} performance on many generation tasks including summarization, question generation, etc. 

% 	\subsection{Other methods}
% 	Based on the similarity between our task and abstractive summarization and story generation (with given topic words), we also apply \textit{Pointer Generator Networks} (``PointerGen'')~\cite{See2017GetTT} and \textit{Multi-scale Fusion Attention} (``Fusion Attn.'')~\cite{Fan2018HierarchicalNS} model respectively for our task. 
%	Moreover, we also investigate a state-of-the-art method for keywords-based sentence sampling from language models, named \textit{CGMH}~\cite{Miao2018CGMHCS}, which is an edit-based lexically-constrained decoding algorithm without training or fine-tuning.

% GPT-2~\cite{radford2019language}, BERT-gen~\cite{Devlin2019},  UniLM~\cite{Dong2019UnifiedLM}, UniLM-v2~\cite{bao2020unilmv2}, BART~\cite{Lewis2019BARTDS}, and T5~\cite{raffel2019exploring} 
% \smallskip
% \noindent
% \textbf{Imposing Commonsense Knowledge.}
% \subsection{Non-autoregressive Generative Models}

%	Thus, the problem is more like an abstractive summarization task.  

% \subsection{Graph-to-Seq Methods}

% \subsection{Constrained Sampling Methods}

%% file: 4_eval.tex
	\section{Evaluation}
	\label{sec:evaluation}

	We first introduce the automatic evaluation metrics, then present main experimental results with manual analysis, and finally introduce the potential application in transferring CommonGen-trained models for other downstream tasks.
% 	We first introduce the setup and automatic metrics, and then we present the results and analysis.
% 	Finally, we show human evaluation results and qualitative analysis. 
	
% 	\subsection{Setup}
% 	We use the proposed \textsc{CommonGen} dataset in two setting: knowledge-agnostic and knowledge-aware.
% 	For the knowledge-agnostic setting, we simply apply the methods in Section~\ref{sec:baseline} while we concatenate rationales and input concept-sets together as the knowledge-aware inputs (``$+omcs$'').

	\subsection{Metrics}
    Following other conventional generation tasks, we use several widely-used automatic metrics to automatically assess the performance, such as \texttt{BLEU}~\cite{Papineni2001BleuAM}, \texttt{ROUGE}~\cite{Lin2004ROUGEAP},  \texttt{METEOR}~\cite{Banerjee2005METEORAA}, which mainly focus on measuring surface similarities.  We  report the concept \texttt{Coverage}, which is the average percentage of input concepts that are present in lemmatizatized outputs.
    
    In addition, we argue that it is more suitable to use evaluation metrics specially design for captioning task, such as \texttt{CIDEr}~\cite{Vedantam2014CIDErCI} and \texttt{SPICE}~\cite{Anderson2016SPICESP}.
    They usually assume system generations and human references use similar concepts, and thus focus on evaluate the associations between mentioned concepts instead of n-gram overlap.
    For example, the \texttt{SPICE} metric uses dependency parse trees as proxy of scene graphs to measure the similarity of scenarios.\footnote{We also tried recent metrics such as BERTScore~\cite{Zhang2020BERTScore}, but we find that they 
    overly focus on lexical semantics instead of dependencies between words, thus resulting low correlation with the manual evaluation results.}

	To estimate \textit{human performance} within each metric, we treat each reference sentence in dev/test data as a ``system prediction''  to be compared with all other references, which is equivalent to compute inter-annotator agreement within each metric. 
	Thus, systems that have better generative ability than average crowd-workers should exceed this.

	\subsection{Experimental Results}
		\smallskip
	\noindent
	\textbf{Automatic Evaluation.} 
	Table~\ref{tab:exp} presents the experimental results
% 	\footnote{Implementation details like hyper-parameters about training/decoding are detailed in the \textit{reproducibility instructions} within the \textbf{submitted code}. The dev results are in \textbf{Appendix}.} 
	 in a variety of metrics.
% 	The order-insensitive method \texttt{MeanPooling-CopyNet} outperforms its order-sensitive counterparts.
% % 	, but such a marginal improvement is not seen in the comparison between ``Trans.'' and ``Set. Trans.''
% 	We assume that for short sequences the order sensitivity does not harm Transformer encoders too much, but positional embeddings are quite necessary to ensure its self-attention mechanism.  
    We can see that all fine-tuned pre-trained models (the lower group) outperform  non-pretrained models (the upper group) with a significant margin.
% 	We find that vanilla Transformer architectures are not outperforming simpler models like bRNN, which probably is because of its complex structure needs more carefully tuning or a specially copying attention in this setting.
% 	the pretrained NLG model UniLM outperforms all other baseline methods by a large margin,
    % The better performance in edit-based Transformer model, LevenTrans, also suggests that, yielding the best performance among models without pre-training. 
    % We argue that the non-autoregressive models with iterative refinement on previously decoded sentences are more promising in our setting, as our inputs can be naturally viewed as incomplete sentence prototypes.
% 	
% 	We find that the pre-trained models outperform the previous baselines by a large margin. 
	This is not surprising because their pretraining objectives, including masked language modeling, word ordering, and text infilling which predicts missing words or text spans, are relevant to our task. 
	On the other hand, we find that the key disadvantage of non-pretrained models with CopyNet still falls in the failure of using all given concepts (i.e., low coverage), which results in worse results.

	Among them, UniLM, BART, and T5 performs the best, which may be due to its inherent sequence-to-sequence pre-training framework. 	
	We found that \texttt{BART} has the best concept coverage, which is probably due to its comprehensive pre-training tasks that aim to recover text with noise.
	The results suggest that further modifying  pre-trained models is a promising direction for generative commonsense.

	\begin{table}[t!]
	\small
		\centering
		\scalebox{0.86
		}{	 \begin{tabular}{@{}c|c|c|c|c|c|c}
			 	\toprule
      & C.Leven & GPT & BERT-G. & UniLM & BART  & T5    \\ \midrule
Hit@1 & 3.2         & 21.5  & 22.3     & 21.0  & \underline{26.3} & \textbf{26.8} \\
Hit@3 & 18.2        & 63.0  & 59.5     & \underline{69.0}  &\underline{69.0} & \textbf{70.3 }\\
Hit@5 & 51.4        & 95.5  & 95.3     & \underline{96.8} & 96.3 & \textbf{97.8} \\
\bottomrule
\end{tabular}
			} 
		\caption{\small{\textbf{Manual Evaluation via Pair-wise Comparisons for Ranking.} Numbers are hit rates (\%) at top 1/3/5.}
% 		\vspace{-1em}
		}
		\label{tab:maneval}
	\end{table}
	\smallskip
% \bigskip
	\noindent
	\textbf{Manual Evaluation.} 
We conduct manual evaluation with a focus on \textit{commonsense plausibility} for comparing the 6 best-performing models in Table~\ref{tab:maneval}.
We ask five graduate students to compare 1,500 pairs of model-generated sentences respectively, for ranking the models within 100 concept-sets that are covered by all the models. 
The final average ranked results are shown in Table~\ref{tab:maneval} and their inter-annotator agreement is 0.85 in \textit{Kendall's rank correlation  coefficient}.

Note that the coverage-weighted hit@1 rate correlates with the \texttt{SPICE} metric the most, i.e.,\textbf{ 0.94} in \textit{Spearman's} $\rho$ for model ranks, CIDEr for 0.91, while METEOR and ROUGE-2 are both 0.88 and BLEU-4 is 0.78.

	\begin{figure}[t!]
		\centering
		\includegraphics[width=1\linewidth]{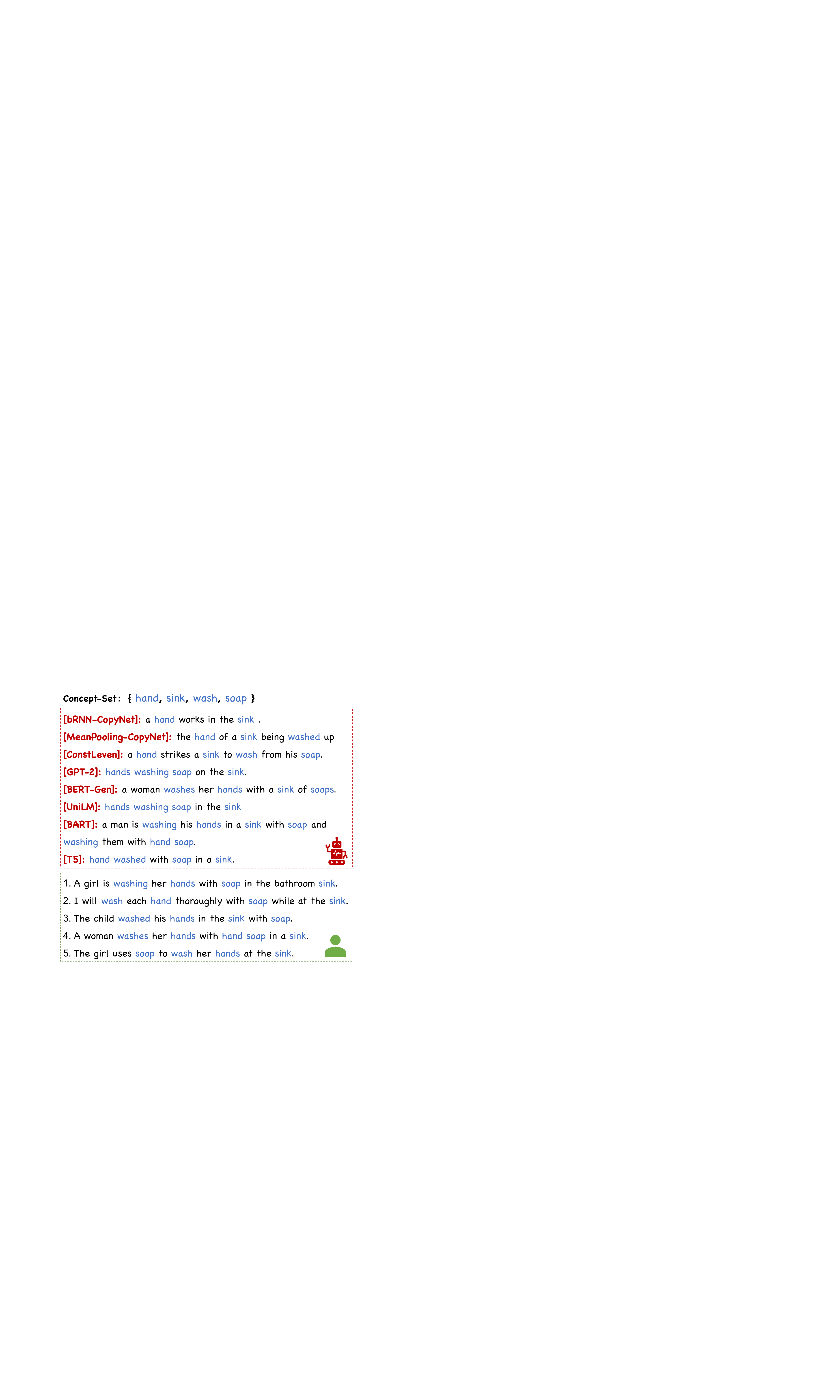}
		\caption{A case study with a concept-set \{\textit{hand}, \textit{sink}, \textit{wash}, \textit{soap}\} for qualitative analysis of machine generations. 
		Human references are collected from AMT.}
		\label{fig:casestudy}
	\end{figure}
	
	\smallskip
	\noindent
	\textbf{Case study.}	
    Fig.~\ref{fig:casestudy} shows the top generations of different models and human references about an input concept-set:  \{\textit{hand}, \textit{sink}, \textit{soup}, \textit{wash}\} (more cases are shown in Fig.~\ref{fig:morecase} in the appendix). 
    % We can find that  can hardly cover all of the given concepts, and the sentences are not quite grammatical. 
    % The LevenTrans model performs better in covering more concepts with correct preposition words but produces repetition of words. 
    We find that non-pretrained seq2seq models (e.g., bRNN, MeanPooling, ConstLeven) can successfully use part of given concepts, while the generated sentences are less meaningful and coherent.
    % % The vanilla \texttt{LevenTrans} model only uses one of the given concepts, although it aims to modeling the edits explicitly and generates syntactically sound sentences.
    % For example, \texttt{bRNN-CopyNet} uses all four concepts with the powerful copy mechanism, but generates nonsensical sentences.
    On the contrary, the outputs of fine-tuned pre-trained language models are significantly more commonsensical.
    % Although they are not equipped with an explicit module for enforcing the use of given concepts,
    Most of them use all given concepts in their outputs.
    ConstLeven tends to make use of frequent patterns to compose a non-sense sentence but uses all concepts.
    GPT-2 and UniLM incorrectly compose the dependency among \textit{hand}, \textit{wash}, and \textit{soap}.
    The phrase `a sink of soaps' in BERT-gen's output makes itself less common.
    BART and T5 generate relatively reasonable scenarios, but both are not as natural as human references; BART's contains repetitive content while T5's lacks a human agent.
    
    \begin{figure}[t!]
		\centering
		\includegraphics[width=0.98\linewidth]{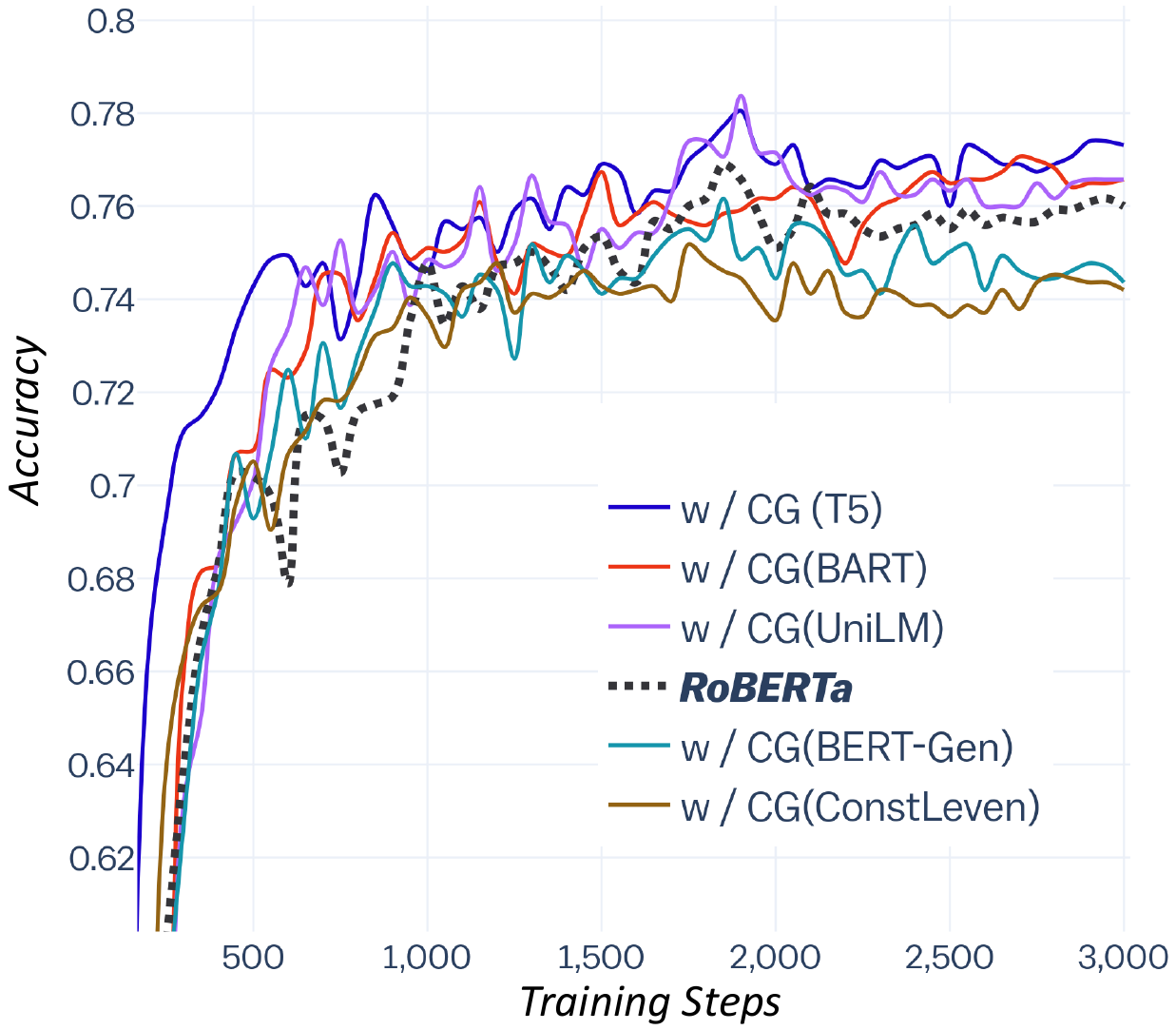}
		\caption{{\textbf{Learning curve for the transferring study.} We use several trained \textsc{CommonGen} (GG) models to generate choice-specific context for the CSQA task.  Detailed numbers are shown in Tab.~\ref{tab:detailtransfer} in the appendix.
% 		The $x$-axis is the number of training steps, while $y$-axis is the accuracy on the dev set.
		}}
		\label{fig:learncurve}
	\end{figure}

\begin{table*}[th!]
	\centering
	\scalebox{0.9
	}{
		\begin{tabular}{@{}c|cc|cc|c|cc|c@{}}
		
\textbf{Model $\backslash$ Metrics}             & \multicolumn{2}{c}{\texttt{ROUGE-2/L}} & \multicolumn{2}{c}{\texttt{BLEU-3/4}}  & \texttt{METEOR} & \texttt{CIDEr} & \texttt{SPICE} &  \texttt{Coverage}  \\ \midrule T5-large+DBA & 16.8    & 36.71   & 27.3   & 18.7   & 25.3   & 8.62  & 24.3  & 83.98    \\
T5-base+DBA  & 15.07   & 34.82   & 24.8   & 16     & 23.5   & 9.31  & 21.3  & 76.81    \\
GPT-2+DBA    & 17.56   & 39.45   & 29.4   & 20.6   & 24.9   & 10.85 & 26.8  & 79.51    \\
BART+DBA     & 18.15   & 37.02   & 28.3   & 19.1   & 25.5   & 9.82  & 25.1  & 84.78   \\  
 
\bottomrule

\end{tabular}
		
	} 
	\caption{Experimental results of models with DBA decoding method on the test set.}
	\label{tab:dbaexp}
\end{table*}

\noindent
 \textbf{Influence of Dynamic Beam Allocation.~}
%  \label{app:dba}
 Considering that all tested models decode sentences with beam searching, one may wonder what if we use a decoding method specially designed for constrained decoding.
 Thus, we employed dynamic beam allocation (DBA)~\cite{post-vilar-2018-fast}.
 The results are shown in Table~\ref{tab:dbaexp}.
 Note that the models are the same as in Table~\ref{tab:exp} while only the decoding method is changed to DBA.
 We can see that all methods are negatively impacted by the decoding method.
 This suggests that for the \textsc{CommonGen} task and pre-trained language models, we may need to focus on knowledge-based decoding or re-ranking as future directions.
	
    \subsection{Transferring CommonGen Models}
    \label{ssec:transfer}
    One may wonder how fine-tuned \textsc{CommonGen} models can benefit commonsense-centric downstream tasks such as Commonsense Question Answering~\cite{Talmor2018CommonsenseQAAQ} (CSQA) with their generative commonsense reasoning ability.
    To this end, we use the models trained with the \textsc{CommonGen} dataset for generating useful context. 
    % For example, a question in CSQA dataset is ``\textit{What do people aim to do at work?}'' with five choices \{complete job(\xmark), {learn from each other} (\cmark), kill animals(\xmark), wear hats (\xmark), talk to each other(\xmark)\}.
    
    We extract the nouns and verbs in questions and all choices respectively, and combine the concepts of the question $q$ and each choice $c_i$ to build five concept-sets. 
    Then, we use these concept-sets as inputs to a trained \textsc{CommonGen} model (e.g., T5) for generating scenario a sentence $g_i$ for each as choice-specific contexts.
    Finally, we prepend the outputs in front of the questions, i.e., ``\textless s\textgreater G: $g_i$ $|$ Q: $q$ \textless /s\textgreater~C: $c_i$ \textless /s\textgreater ''. Note that the  state-of-the-art RoBERTa-based models for CSQA uses the same form without ``G: $g_i|$'' in fine-tuning.

    We show the learning-efficiency curve in Fig.~\ref{fig:learncurve}, where $y$ is the accuracy on the official dev set and $x$ is the number of training steps.
    The details of the experiments are shown in the appendix.
    % Batch-size is 16 and 3k training steps are about 5 epochs. We use the same hyper-parameters which are searched over the baseline RoBERTa-Large model for all experiments here. This suggests that further searching over CommonGen-enhanced models may have even better performance.
    
    We highlight the performance of original RoBERTa-Large as the baseline.
    We find that some CommonGen models further improves the performance by a large margin, e.g., $76.9 \xrightarrow{\texttt{UniLM}} 78.4$ and they converge at better accuracy in the end. Note that BERT-gen and ConstLeven cause negative transfer due to the low quality of generated context. 
    Particularly, we find that the context generated by the T5-based CommonGen model (CG-T5) helps speed up training about 2 times, if we look at 550th steps of CG-T5 (74.85\%) and 1,250th steps of original RoBERTa (74.77\%).

    Through manual analysis, we find that the successful \textsc{CommonGen} models can generate more reasonable and natural sentence for correct choices while noisy sentences for wrong choices.
    For example with CG (T5), $q$=``\textit{What do \underline{people} \underline{aim} to do at \underline{work}?}'',  $c_i$=`\underline{complete} \underline{job}' (\cmark) with $g_i$=``\textit{people work to complete a job aimed at achieving a certain goal.}''; 
    $c_j$=`\underline{wear} \underline{hats}' (\xmark) $g_j$=``\textit{people wearing hats aim their guns at each other while working on a construction site.}''
	The used question concepts and choice concepts are {underlined}.

%% file: 5_related.tex
\section{Related Work}
	\label{sec:relatedwork}
	\noindent
	\textbf{Commonsense benchmark datasets.~}
% 	Machine common sense (MCS) has long been considered as one of the most significant area in artificial intelligence.
%	Recent advances in natural language processing and machine learning have shown that deep neural networks can effectively learn to understand images and language for many context-restricted reasoning tasks such as machine reading comprehension~\cite{Rajpurkar2016SQuAD10}.
%	However, they can be far from human performance in some commonsense reasoning tasks.
%	
	There are many emerging datasets for testing machine commonsense from different angles,
	such as commonsense extraction~\cite{Xu2018AutomaticEO,Li2016CommonsenseKB}, next situation prediction ({SWAG}~\cite{Zellers2018SWAGAL}, CODAH~\cite{Chen2019CODAHAA}, HellaSWAG~\cite{Zellers2019HellaSwagCA}), cultural and social understanding~\cite{Lin2018MiningCD, sap2018atomic, sap-etal-2019-social}, visual scene comprehension~\cite{Zellers2019FromRT}, and general commonsense question answering~\cite{Talmor2018CommonsenseQAAQ, huang-etal-2019-cosmos, wang-etal-2019-make,wang-etal-2020-semeval}. 
	   However, the success of fine-tuning pre-trained language models for these tasks does not necessarily mean machines can produce novel assumptions in a more open, realistic, generative setting.
	We see \textsc{CommonGen} as a novel, complementary  commonsense reasoning benchmark task for advancing machine commonsense in NLG.

	\smallskip
	\noindent
	\textbf{Constrained Text Generation.~}
% 	\yuchen{Add Yue Zhang's works on word ordering}
	Constrained text generation aims to decode sentences with expected attributes such as sentiment~\cite{Luo2019TowardsFT, Hu2017TowardCG}, tense~\cite{Hu2017TowardCG}, template~\cite{Zhu2019TextI, j-kurisinkel-chen-2019-set}, style~\cite{fu2018style, Luo2019ADR, Li2018DeleteRG}, topics~\cite{Feng2018TopictoEssayGW}, etc.
	Two related scenarios with our task is lexically constrained decoding and word ordering~\cite{Zhang2015DiscriminativeSW, Hasler2018NeuralMT, Dinu2019TrainingNM, Hokamp2017LexicallyCD, puduppully-etal-2017-transition,Miao2018CGMHCS}. 
	However, they are not easily adopted by the recent pre-trained language models and thus not directly useful for our task.
% 	One recent work in this line is the CGMH~\cite{Miao2018CGMHCS} method, which aims to sample sentences with an \textit{ordered} sequence of keywords from language models but cannot be fine-tuned and adopted in our case. 
	Topical story generation~\cite{Fan2018HierarchicalNS, yao2019plan} is also a related direction, while it targets generating longer, creative stories around the given topics, making it hard to directly adopt them to our task.
	Additionally, the \textsc{CommonGen} task brings some more challenges mentioned in Section~\ref{sec:formualtion}.
	Prior constrained generation methods cannot address these issues together in a unified model.
% 	, 
% 	and we expect \textsc{CommonGen} to be also a benchmark dataset for future works in this direction. 
	
	\smallskip
	\noindent
	\textbf{Incorporating Commonsense for NLG.~}
% 	Though pre-trained language models have captured commonsense to some extent, how 
	There are a few recent works that incorporate commonsense knowledge in language generation tasks such as essay generation~\cite{Guan2018StoryEG, Yang2019EnhancingTG}, image captioning~\cite{DBLP:conf/cvpr/LuYBP18}, video storytelling~\cite{Yang2019KnowledgeableSA}, and conversational systems~\cite{Zhang2019ConversationGW}.
	These works suggest that generative commonsense reasoning has a great potential to benefit downstream applications.
    Our proposed \textsc{CommonGen}, to the best of our knowledge, is the very first constrained sentence generation dataset for assessing and conferring generative machine commonsense and we hope it can benefit such applications.
    Our transferring study in Sec.~\ref{ssec:transfer} also shows the potential benefits of CommonGen-generated contexts.
% 	We instead mainly investigate sequence-to-sequence architectures, especially models that are based on editing operations and non-autoregressive. 
% 	Pre-trained seq2seq generation models like UniLM~\cite{Dong2019UnifiedLM} and BRAT~\cite{Lewis2019BARTDS} are usually initialized with pre-trained language encoder and then further fine-tuned with multiple NLG tasks.
% 	The UniLM archives the best performance on our proposed \textsc{CommonGen} task, while being far from human-level performance and hardly interpretable.

%% file: 6_conclu.tex
	\section{Conclusion}
	\label{sec:conclusion} 
	Our major contribution in this paper are threefold:
	\begin{itemize}
	    \item  we present \textsc{CommonGen}, a novel constrained text generation task for generative commonsense reasoning, with a large dataset;
	    \item we carefully analyze the inherent challenges of the proposed task, i.e.,  a) relational reasoning with latent commonsense knowledge, and b) compositional generalization.
	    \item our extensive experiments systematically examine recent pre-trained language generation models (e.g., UniLM, BART, T5) on the task , and find that their performance is still far from humans, generating grammatically sound yet realistically implausible sentences.
	\end{itemize}
% 	The dataset contains about 30k concept-sets as input with a diverse concept vocabulary.
% 	The task is inherently challenging as the models need to understand and use background knowledge that underlies given concepts while maintaining compositaional generalization ability. 
% 	Additionally the task requires complex relational commonsense reasoning to generate an expected sentence, a bottleneck in many current state of the art natural language systems.
% 	Most widely used baseline methods cannot produce desirable sentences, even when given a set of relevant commonsense facts as additional context.
% 	Even the best baseline model UniLM, a BERT-based NLG model, still performs significantly worse than human performance.
    Our study points to interesting future research directions on modeling commonsense knowledge in language generation process, towards conferring machines with generative commonsense reasoning ability. 
	We hope \textsc{CommonGen} would also benefit downstream NLG applications such as conversational systems and storytelling models.
% 	We describe a data collection process for crowd-sourcing high quality human-written sentences at scale. 

% 	For the future research, we believe the following directions are highly valuable to explore:
% 	1) specially designed metrics for automatic evaluation that focus on commonsense plausibility; 2) better mechanisms for retrieving and imposing useful commonsense knowledge into sentence generation processes; 3) explicitly modeling keyword-centric edits (e.g. insertion, deletion, morphological changes) such that relevant commonsense knowledge can be well utilized.
% 	We also believe that models performed well on \textsc{CommonGen} can be easily transferred to other commonsense-required reasoning tasks with few annotations,  including image/video captioning, visual question answering, and discriminative multi-choice commonsense question answering. 

%% file: appendix.tex
\appendix
\section{Supplementary Figures and Tables}
\begin{figure*}[t!]
		\centering
		\includegraphics[width=1\linewidth]{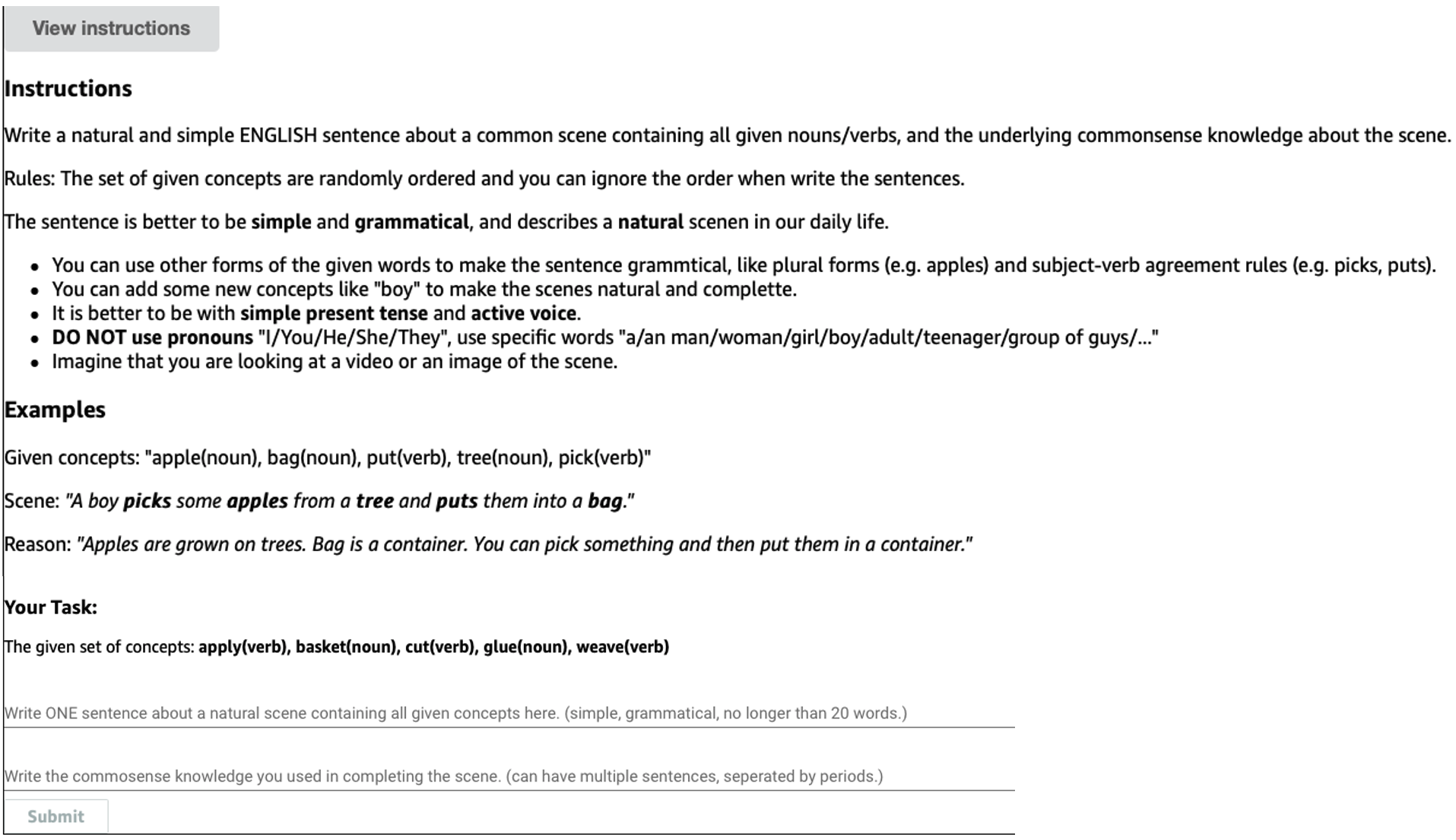}
		\caption{Our annotation interface on the AMT platform. The upper part is the instruction for the annotators and we provide an example for them. Note that we give the part-of-speech hints (from the captain corpora) to boost the speed of annotation, but we do not remove sentences with wrong part-of-speech as long as they also make sense.}
		\label{fig:amt}
	\end{figure*}

\begin{figure*}[t!]
		\centering
		\includegraphics[width=1\linewidth]{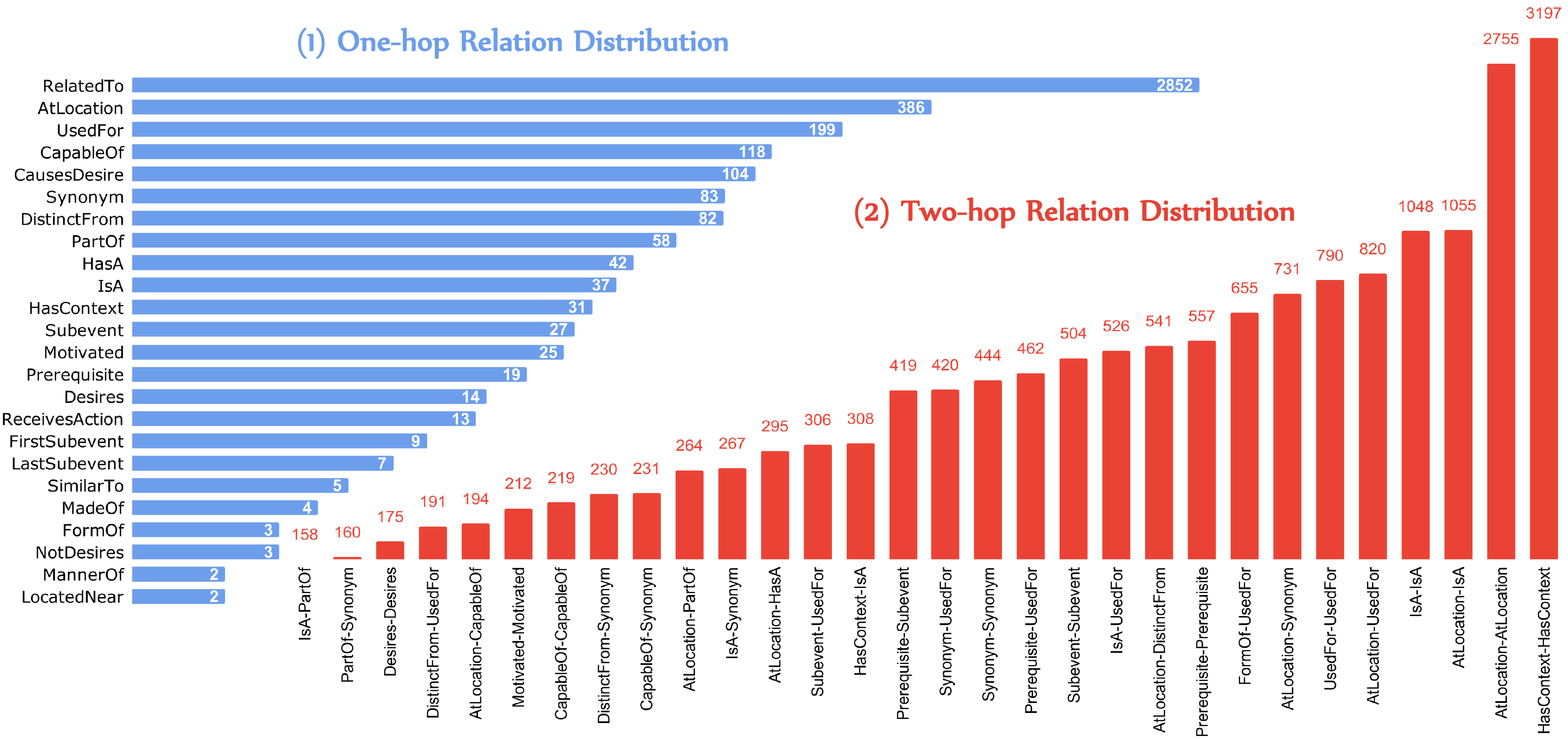}
		\caption{One/two-hop relation frequency in the \textsc{CommonGen} dev.\&test sets on ConceptNet.}
		\label{fig:relationdist}
	\end{figure*}

\begin{figure*}[t!]
		\centering
		\includegraphics[width=1.0\linewidth]{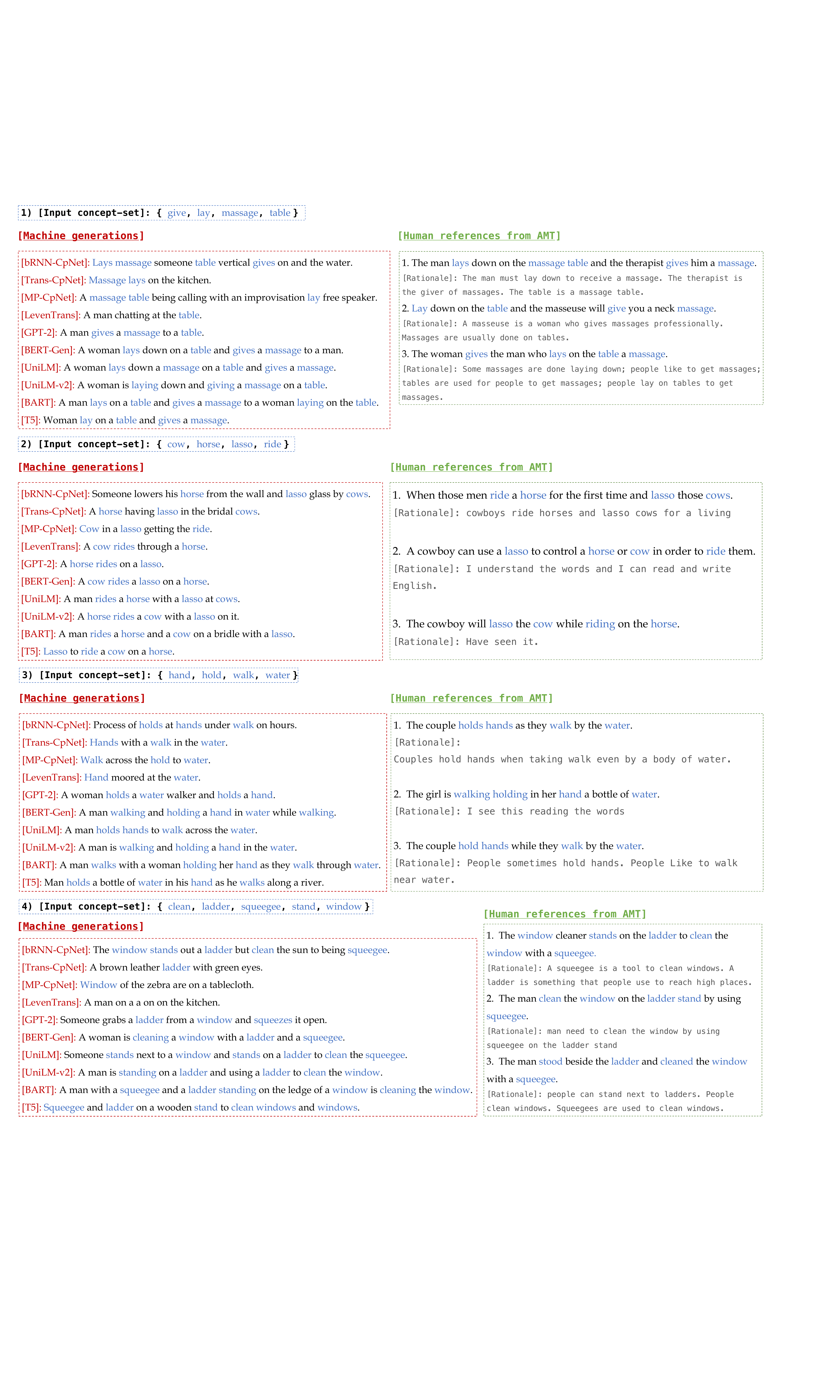}
		\caption{Four cases for qualitative analysis of machine generations. 
		References are collected from AMT crowd-workers and they are required to provide rationales. Note that the third one is a positive case showing that some models can successfully generate reasonable scenarios. However, most models perform poorly on the other cases.}
		\label{fig:morecase}
	\end{figure*}

We include additional figures and tables that we mentioned in the main content here.
\begin{itemize}
    \item Figure~\ref{fig:relationdist} shows the detailed \textbf{distribution of the commonsense relations} between given concepts, the summary of which was shown in Table~\ref{tab:cate} of the main content.
    \item  Figure~\ref{fig:morecase} presents 4 more \textbf{case studies} with human rationales which we asked our crowd workers to provide.  
    \item  Figure~\ref{fig:amt} shows instructions and AMT interface for crowd-sourcing human references.
    \item Table~\ref{tab:devexp} shows the model performances on the dev set of \textsc{CommonGen}, as a reference for future development.
    \item Table~\ref{tab:detailtransfer} is the full results of the learning curve in Figure~\ref{fig:casestudy}. We highlight the highest checkpoints and the speed-up by the CG-T5, which are discussed in Section~\ref{ssec:transfer}.
\end{itemize}

\section{Experimental Details}
\begin{figure}[t!]
	\centering
	\includegraphics[width=1\linewidth]{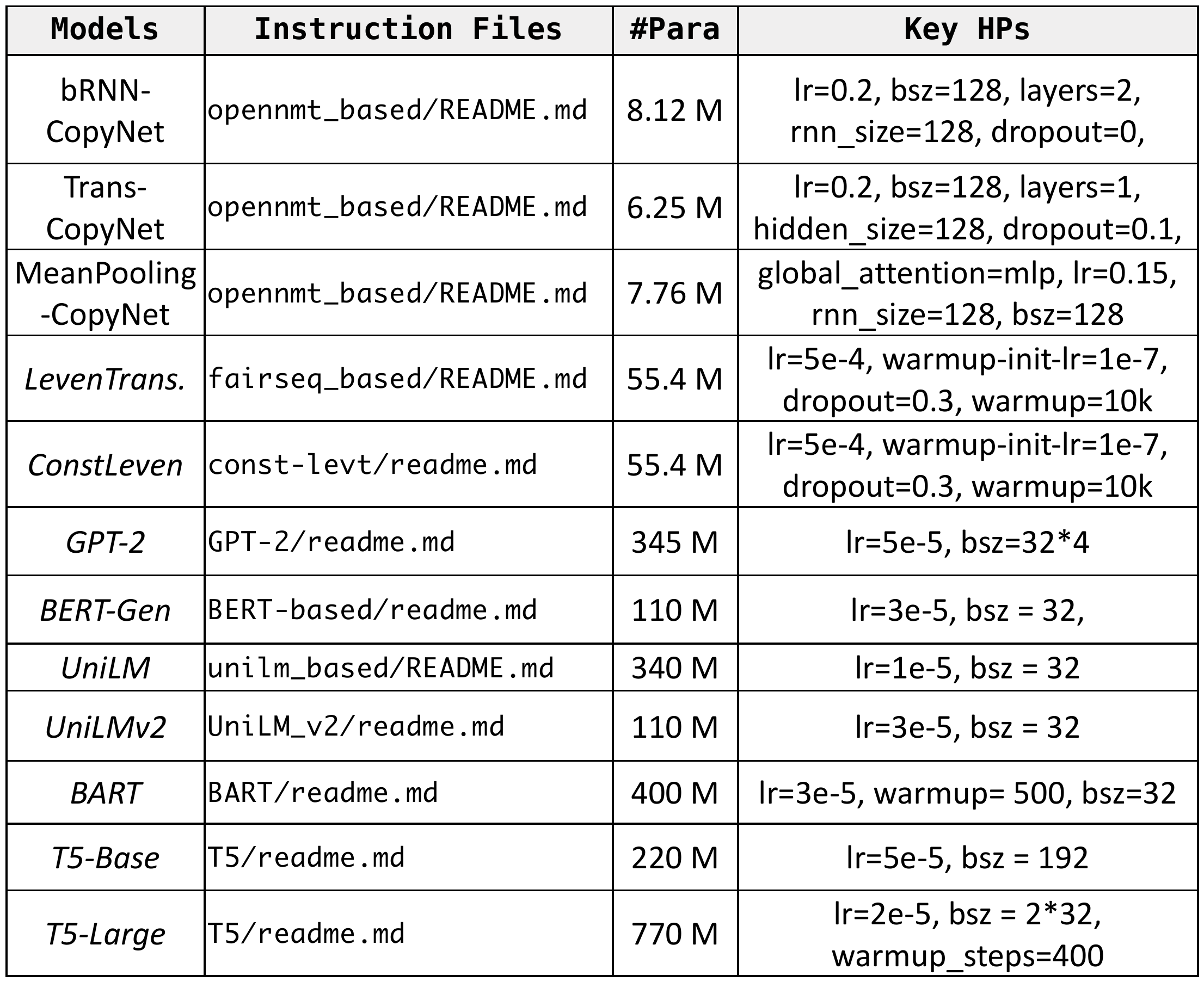}
	\captionof{table}{The paths to the instruction files in our submitted code zip file (under the `\textit{methods/}' folder), and their numbers of parameters and key hyper-parameters.}
\label{tab:model_details}
\end{figure}

\textbf{Main experiments.}\quad 
We present some implementation details in training and testing the baseline models in Table~\ref{tab:model_details}.
The detailed instructions for installing dependencies and all necessary training command-lines are shown in the instruction `\textbf{readme.md}' files.
The number of trainable model parameters are directly induced from either output of the frameworks or the original papers.
We show some key hyper-parameters that we manually tuned on top of the development set.

All key hyper-parameters were initialized by the default values as suggested by the original authors of the frameworks.
The bound of our manual tuning is done by iterating the magnitudes or the neighboring choices, for example, the learning rates (`lr') of the last seven models  are selected from $\{1e-3,\dots,1e-4, \dots, 1e-5\}$.
Then, similarly, the batch size (bsz) is first maximized by making full use of the GPU memory.
Note that the first three models are implemented with the OpenNMT-py framework~\footnote{\url{https://github.com/OpenNMT/OpenNMT-py}}.
The \texttt{LevenTrans}\footnote{\url{https://github.com/pytorch/fairseq/blob/master/examples/nonautoregressive_translation/README.md}}, \texttt{ConstLeven}\footnote{\url{https://github.com/raymondhs/constrained-levt}}, and \texttt{BART}\footnote{\url{https://github.com/pytorch/fairseq/tree/master/examples/bart}} are adopted by the official authors' release.
The \texttt{BERT-gen}, \texttt{UniLM}, \texttt{UniLMv2} are all based on their official source code\footnote{\url{https://github.com/microsoft/unilm}}.
The \texttt{GPT-2} and \texttt{T5} are both adopted by the \texttt{huggingface} transformers\footnote{\url{https://github.com/huggingface/transformers}} framework~\cite{Wolf2019HuggingFacesTS}.
All models use beam searching as their decoding algorithms and beam-size are mostly 5, which is selected from $\{5,10,20\}$.
All our models were trained on Quadro RTX 6000 GPUs.
The training time of X-CopyNet and LevenTrans models are less than 12 hours with a single GPU.
The second group of models are trained between 12 and 24 hours, expect for T5-large, which we used 3 GPUs and fine-tuned about 48 hours.
\textit{Note that all the above methods are \underline{self-contained} in our submitted code as long as users follow the associated readme instructions.}

\textbf{Transferring study experiments.}\quad 
 We use the same hyper-parameters which are searched over the baseline RoBERTa-Large model for these experiments. 
 The best hyper-parameter\footnote{We follow the hps selected by 100 trials of tuning in \url{https://github.com/pytorch/fairseq/tree/master/examples/roberta/commonsense_qa}.} of RoBERTa-Large for \texttt{CommonsenseQA}\footnote{\url{https://www.tau-nlp.org/commonsenseqa}}: 
 \begin{itemize}
     \item batch size = 16, learning rate = 1e-5,
     \item maximum updates = 3,000 ($\sim$5 epochs) \item warmup steps=150, dropout rate=0.1
     \item weight decay = 0.01, adam\_epsilon = 1e-6
 \end{itemize}

% \begin{table*}[th!]
% 	\centering
% 	\scalebox{0.9
% 	}{
% 		\begin{tabular}{@{}c|cc|cc|c|cc|c@{}}
		
% \textbf{Model $\backslash$ Metrics}             & \multicolumn{2}{c}{\texttt{ROUGE-2/L}} & \multicolumn{2}{c}{\texttt{BLEU-3/4}}  & \texttt{METEOR} & \texttt{CIDEr} & \texttt{SPICE} &  \texttt{Coverage}  \\ \midrule T5-large+DBA & 16.8    & 36.71   & 27.3   & 18.7   & 25.3   & 8.62  & 24.3  & 83.98    \\
% T5-base+DBA  & 15.07   & 34.82   & 24.8   & 16     & 23.5   & 9.31  & 21.3  & 76.81    \\
% GPT-2+DBA    & 17.56   & 39.45   & 29.4   & 20.6   & 24.9   & 10.85 & 26.8  & 79.51    \\
% BART+DBA     & 18.15   & 37.02   & 28.3   & 19.1   & 25.5   & 9.82  & 25.1  & 84.78   \\  
 
% \bottomrule

% \end{tabular}
		
% 	} 
% 	\caption{Experimental results of models with DBA decoding method on the test set.}
% 	\label{tab:dbaexp}
% \end{table*} 

\begin{table*}[th!]
	\centering
	\scalebox{0.85
	}{
		\begin{tabular}{@{}c|cc|cc|c|cc|c@{}}
		
\textbf{Model $\backslash$ Metrics}             & \multicolumn{2}{c}{\texttt{ROUGE-2/L}} & \multicolumn{2}{c}{\texttt{BLEU-3/4}}  & \texttt{METEOR} & \texttt{CIDEr} & \texttt{SPICE} &  \texttt{Coverage}  \\ \midrule 
bRNN-CopyNet~\cite{gu-etal-2016-incorporating} & 9.23 & 30.57 & 13.60 & 7.80 & 17.40 & 6.04 & 16.90 &  58.95 \\ 
% bRNN w/ OMCS & 4.15 & 21.74 & 7.70 & 2.80 & 15.10 & 5.22 & 14.30 & 37.20 & 50.67 \\ 
% bRNN w/ order & 4.28 & 30.44 & 8.40 & 3.60 & 18.20 & 6.68 & 14.70 & 53.20 & 67.16 & 7.19 \\  
% Set-Trans.~\cite{Vaswani2017AttentionIA} & 1.59 & 12.96 & 3.20 & 1.20 & 8.60 & 2.02 & 7.00 & 17.71 & 24.10 \\ 
Trans-CopyNet & 11.08 & 32.57 & 17.20 & 10.60 & 18.80 & 7.02 & 18.00 & 62.16 \\
MeanPooling-CopyNet & 11.36 & 34.63 & 14.80 & 8.90 & 19.20 & 7.17 & 20.20 & 68.32 \\ 
% Trans. w/ OMCS & 1.73 & 12.81 & 4.20 & 1.40 & 9.20 & 2.48 & 8.10 & 14.93 & 20.87  \\
% Trans. w/ order & 3.31 & 18.39 & 5.90 & 2.70 & 12.50 & 3.74 & 9.70 & 32.27 & 38.71 & 3.72 \\ 
LevenTrans.~\cite{Gu2019LevenshteinT} & 12.22 & 35.42 & 23.10 & 15.00 & 22.10 & 8.94 & 21.40 & 71.83 \\ 
ConstLeven.~\cite{Susanto2020LexicallyCN} & 13.47 & 35.19 & 21.30 & 12.30 & 25.00 & 11.06 & 23.20 & 96.87 \\ 
\midrule   
GPT-2~\cite{radford2019language} & 17.74 & 41.24 & 32.70 & 23.30 & 27.50 & 13.26 & 27.60 & 85.46  \\  
BERT-Gen~\cite{bao2020unilmv2}  & 18.73 & 42.36 & 33.00 & 23.70 & 29.10 & 13.34 & 28.70 & 91.71  \\
UniLM~\cite{Dong2019UnifiedLM} & 21.68 & \textbf{45.66} & \underline{40.40} & \underline{30.40} & \textbf{31.00} & \underline{15.72} & \underline{31.40} & 92.41 \\ 
% UniLM w/ OMCS & 20.77 & 41.25 & 36.40 & 25.70 & 29.30 & 14.08 & 29.20 & 86.32 & 89.42 \\ 
UniLM-v2~\cite{bao2020unilmv2} & 19.24 & 43.01 & 33.40 & 24.20 & 29.20 & 13.65  & 29.30 & 93.57  \\  
BART~\cite{Lewis2019BARTDS} & \textbf{22.13} & 43.02 & 37.00 & 27.50 & \textbf{31.00} & 14.12 & 30.00  & \textbf{97.56}  \\  
T5-Base~\cite{raffel2019exploring} & 15.33 & 36.20 &  28.10 &  18.00 & 24.60 & 9.73 & 23.40 &  83.77  \\
 T5-Large~\cite{raffel2019exploring} & \underline{21.98} & \underline{44.41} &  \textbf{40.80} &  \textbf{30.60} & \textbf{31.00} & \textbf{15.84} & \textbf{31.80} &  \underline{97.04}  \\  

% \midrule
% % \rowcolor{LightCyan} 
% Human Performance & 48.88 & 63.79 & 48.20 & 44.90 & 36.20 & 43.53 & 63.50 & 99.31  \\ 
\bottomrule

\end{tabular}
		
	} 
	\caption{Experimental results of different baseline methods on the \textsc{CommonGen} dev set. The first group of models are non-pretrained models, while the second group is large pretrained models that we have fine-tuned. The best models are \textbf{bold} and second best ones are \underline{underlined} within each metric.}
	\label{tab:devexp}
\end{table*}

 We tried 10 random seeds and use the best one (42).
 Then, we follow the steps described in Sec.~\ref{ssec:transfer} to run other CG-enhanced models with the same hps.
 This suggests that further searching for them may have even better performance.

%  \section{Influence of Dynamic Beam Allocation}
%  \label{app:dba}
%  Considering that all tested models decode sentences with beam searching, one may wonder what if we use a decoding method specially designed for constrained decoding.
%  Thus, we employed dynamic beam allocation (DBA)~\cite{post-vilar-2018-fast}.
%  The results are shown in Table~\ref{tab:dbaexp}.
%  Note that the models are the same as in Table~\ref{tab:exp} while only the decoding method is changed to DBA.
%  We can see that all methods are negatively impacted by the decoding method.
%  This suggests that for the \textsc{CommonGen} task and pre-trained language models, we may need to focus on knowledge-based decoding or re-ranking as future directions.

\begin{table*}[th!]
	\centering
	\scalebox{0.8
	}{
\begin{tabular}{c|cccccc@{}}
\toprule
Training Steps & RoBERTa-Large & w/CG(BART) & w/CG(T5) & w/CG(UniLM) & w/CG(BERT-Gen) & w/CG(ConstLeven) \\
\midrule
50             & 0.2252  & 0.1884     & 0.2506   & 0.2244      & 0.2007         & 0.2162            \\
100            & 0.3088  & 0.2703     & 0.3587   & 0.3153      & 0.2924         & 0.2809            \\
150            & 0.5053  & 0.2973     & 0.5643   & 0.1851      & 0.3391         & 0.3653            \\
200            & 0.5717  & 0.4439     & 0.6650   & 0.3833      & 0.5274         & 0.5324            \\
250            & 0.6020  & 0.5242     & 0.6937   & 0.5348      & 0.5839         & 0.6396            \\
300            & 0.6388  & 0.6601     & 0.7117   & 0.6323      & 0.6274         & 0.6634            \\
350            & 0.6675  & 0.6814     & 0.7150   & 0.6503      & 0.6626         & 0.6740            \\
400            & 0.6830  & 0.6830     & 0.7215   & 0.6847      & 0.6781         & 0.6773            \\
450            & 0.7027  & 0.7068     & 0.7338   & 0.6921      & 0.7068         & 0.6962            \\
500            & 0.7019  & 0.7076     & 0.7428   & 0.7011      & 0.6929         & 0.7052            \\
550            & 0.6978  & 0.7248     & \underline{0.7486}   & 0.7256      & 0.7068         & 0.6904            \\
600            & 0.6790  & 0.7232     & 0.7494   & 0.7338      & 0.7248         & 0.7068            \\
650            & 0.7150  & 0.7289     & 0.7428   & 0.7469      & 0.7101         & 0.7117            \\
700            & 0.7142  & 0.7453     & 0.7477   & 0.7387      & 0.7305         & 0.7183            \\
750            & 0.7027  & 0.7453     & 0.7314   & 0.7527      & 0.7166         & 0.7183            \\
800            & 0.7158  & 0.7355     & 0.7437   & 0.7371      & 0.7281         & 0.7240            \\
850            & 0.7174  & 0.7445     & 0.7625   & 0.7420      & 0.7379         & 0.7322            \\
900            & 0.7191  & 0.7543     & 0.7559   & 0.7502      & 0.7477         & 0.7338            \\
950            & 0.7355  & 0.7486     & 0.7477   & 0.7387      & 0.7428         & 0.7404            \\
1000           & 0.7477  & 0.7510     & 0.7461   & 0.7486      & 0.7428         & 0.7363            \\
1050           & 0.7346  & 0.7502     & 0.7568   & 0.7469      & 0.7412         & 0.7297            \\
1100           & \underline{0.7428}  & 0.7527     & 0.7551   & 0.7494      & 0.7363         & 0.7420            \\
1150           & 0.7379  & 0.7609     & 0.7576   & 0.7641      & 0.7453         & 0.7437            \\
1200           & 0.7469  & 0.7477     & 0.7502   & 0.7461      & 0.7420         & 0.7477            \\
1250           & 0.7477  & 0.7412     & 0.7592   & 0.7518      & 0.7273         & 0.7371            \\
1300           & 0.7502  & 0.7518     & 0.7617   & 0.7666      & 0.7518         & 0.7412            \\
1350           & 0.7469  & 0.7502     & 0.7551   & 0.7568      & 0.7437         & 0.7404            \\
1400           & 0.7420  & 0.7494     & 0.7641   & 0.7559      & 0.7494         & 0.7428            \\
1450           & 0.7510  & 0.7584     & 0.7625   & 0.7461      & 0.7461         & 0.7461            \\
1500           & 0.7535  & 0.7674     & 0.7690   & 0.7551      & 0.7412         & 0.7428            \\
1550           & 0.7461  & 0.7559     & 0.7674   & 0.7510      & 0.7445         & 0.7412            \\
1600           & 0.7437  & 0.7584     & 0.7584   & 0.7543      & 0.7445         & 0.7420            \\
1650           & 0.7568  & 0.7609     & 0.7633   & 0.7543      & 0.7494         & 0.7428            \\
1700           & 0.7551  & 0.7584     & 0.7633   & 0.7625      & 0.7535         & 0.7396            \\
1750           & 0.7600  & 0.7568     & 0.7699   & 0.7740      & 0.7551         & \textbf{0.7518}            \\
1800           & 0.7617  & 0.7559     & 0.7731   & 0.7740      & 0.7527         & 0.7486            \\
1850           & \textbf{0.7690}  & 0.7584     & 0.7772   & 0.7707      & \textbf{0.7617}         & 0.7461            \\
1900           & 0.7658  & 0.7592     & \textbf{0.7805}   & \textbf{0.7838}      & 0.7486         & 0.7445            \\
1950           & 0.7584  & 0.7617     & 0.7715   & 0.7715      & 0.7510         & 0.7396            \\
2000           & 0.7510  & 0.7617     & 0.7690   & 0.7715      & 0.7445         & 0.7355            \\
2050           & 0.7551  & 0.7641     & 0.7731   & 0.7649      & 0.7559         & 0.7477            \\
2100           & 0.7641  & 0.7617     & 0.7641   & 0.7625      & 0.7559         & 0.7412            \\
2150           & 0.7584  & 0.7543     & 0.7658   & 0.7641      & 0.7527         & 0.7461            \\
2200           & 0.7584  & 0.7477     & 0.7649   & 0.7633      & 0.7453         & 0.7371            \\
2250           & 0.7551  & 0.7559     & 0.7641   & 0.7609      & 0.7461         & 0.7363            \\
2300           & 0.7535  & 0.7600     & 0.7699   & 0.7674      & 0.7412         & 0.7420            \\
2350           & 0.7551  & 0.7617     & 0.7682   & 0.7625      & 0.7502         & 0.7412            \\
2400           & 0.7559  & 0.7649     & 0.7699   & 0.7625      & 0.7559         & 0.7387            \\
2450           & 0.7584  & 0.7674     & 0.7707   & 0.7658      & 0.7477         & 0.7387            \\
2500           & 0.7551  & 0.7649     & 0.7600   & 0.7633      & 0.7502         & 0.7363            \\
2550           & 0.7592  & 0.7658     & 0.7731   & 0.7658      & 0.7518         & 0.7387            \\
2600           & 0.7559  & 0.7658     & 0.7715   & 0.7600      & 0.7420         & 0.7371            \\
2650           & 0.7576  & 0.7674     & 0.7690   & 0.7600      & 0.7494         & 0.7420            \\
2700           & 0.7568  & \textbf{0.7707}    & 0.7690   & 0.7600      & 0.7461         & 0.7379            \\
2750           & 0.7568  & 0.7699     & 0.7674   & 0.7649      & 0.7445         & 0.7437            \\
2800           & 0.7592  & 0.7682     & 0.7690   & 0.7617      & 0.7445         & 0.7453            \\
2850           & 0.7592  & 0.7641     & 0.7707   & 0.7649      & 0.7461         & 0.7445            \\
2900           & 0.7609  & 0.7649     & 0.7740   & 0.7658      & 0.7477         & 0.7437            \\
2950           & 0.7617  & 0.7649     & 0.7740   & 0.7658      & 0.7469         & 0.7437            \\
3000           & 0.7600  & 0.7658     & 0.7731   & 0.7658      & 0.7437         & 0.7420     \\   
\bottomrule
\end{tabular}
		
	} 
	\caption{Experimental results of the transferring study on CommonsenseQA dev set.}
	\label{tab:detailtransfer}
\end{table*}